\documentclass{WileyMSP-template}
\usepackage[dvipsnames]{xcolor}
\usepackage{amsmath}
\usepackage{subfigure}
\usepackage{makecell}
\usepackage{adjustbox}
\usepackage{multirow}
\usepackage[flushleft]{threeparttable}

\newcommand{\Skip}[1]{}

\begin{document}

\pagestyle{fancy}

\emergencystretch 3em

\title{A Large-Depth-Range Layer-Based Hologram Dataset for Machine Learning-Based 3D Computer-Generated Holography}




\maketitle

\author{Jaehong Lee}
\author{You Chan No}
\author{YoungWoo Kim}
\author{Duksu Kim*}


\begin{affiliations}
Jaehong Lee, You Chan No, YoungWoo Kim, Prof. Duksu Kim \\
Korea University of Technology \& Education (KOREATECH) \\
Cheonan-si, Chungcheongnam-do, Republic of Korea \\ 
Email Address: bluekds@koreatech.ac.kr \\
\end{affiliations}



\newcommand{\addcite}{\textcolor{red}{[?]}}
\newcommand{\JH}[1]{\textcolor{OliveGreen}{\bf JH: #1}}

\newcommand{\datasetname}[0]{\textit{KOREATECH-CGH}}
\newcommand{\MIT}[0]{\textit{MIT-CGH-4K}}
\newcommand{\PSNRFIP}[0]{PSNR\textsubscript{FIP}(dB)}
\newcommand{\SSIMFIP}[0]{SSIM\textsubscript{FIP}}

\begin{abstract}
Machine learning-based computer-generated holography (ML-CGH) has advanced rapidly in recent years, yet progress is constrained by the limited availability of high-quality, large-scale hologram datasets. To address this, we present \datasetname, a publicly available dataset comprising 6,000 pairs of RGB-D images and complex holograms across resolutions ranging from $256 \times 256$ to $2048 \times 2048$, with depth ranges extending to the theoretical limits of the angular spectrum method for wide 3D scene coverage.
To improve hologram quality at large depth ranges, we introduce amplitude projection, a post-processing technique that replaces amplitude components of hologram wavefields at each depth layer while preserving phase. This approach enhances reconstruction fidelity, achieving 27.01 dB PSNR and 0.87 SSIM, surpassing a recent optimized silhouette-masking layer-based method by 2.03 dB and 0.04 SSIM, respectively.
We further validate the utility of \datasetname~through experiments on hologram generation and super-resolution using state-of-the-art ML models, confirming its applicability for training and evaluating next-generation ML-CGH systems.
\end{abstract}
\section{Introduction}

Computer-Generated Holography (CGH) simulates the propagation of light to synthesize wavefronts that represent three-dimensional scenes.  
By leveraging the principles of wave optics, CGH can render photorealistic depth cues and occlusion effects, making it a promising candidate for next-generation 3D displays~\cite{2019ShimobabaRN622, 2020KyojiRN282}.  
However, the computational cost of generating holograms using physical models, such as the angular spectrum method or Fresnel propagation, has long hindered their real-time application in consumer devices.  
Although recent CGH algorithms utilize hardware accelerators such as GPUs~\cite{201843baba, ichikawa2012full, 2018SatoRN97, 2021BlinderRN241, 2023LeeRN721, 2024LeeRN501}, FPGAs~\cite{2015pangRN817,2019KimRN818, 2022RN821HORN}, and ASICs~\cite{2017SeoRN816} to reduce computational complexity and enhance scalability, the inherently intensive nature of wave-optics calculations continues to pose a challenge for deployment in commercial systems.

Recent advances in machine learning (ML) have significantly accelerated hologram generation.
ML-based CGH (ML-CGH) approaches, such as Tensor Holography~\cite{2021ShiRN222}, have demonstrated the capability to generate high-quality, full-color holograms in real time, even on mobile devices.
These methods leverage deep learning to bypass explicit wave simulations, learning to approximate the physical behavior of light through data-driven models.
ML-CGH offers advantages not only in hologram generation but also in reconstruction.
Phase optimization using stochastic gradient descent and camera-in-the-loop (CITL) calibration techniques~\cite{2020PengRN257,2021choiRN258}, applied with phase-only spatial light modulators (SLMs), yield clean reconstructed images.
These approaches are inherently trainable and benefit from modern deep-learning frameworks.
As a result, ML-CGH has emerged as a vital subfield, enabling real-time 3D holography for applications such as augmented reality (AR), virtual reality (VR), and more.
As with other ML domains, the quality and diversity of training datasets play a critical role in determining the generalization ability and performance of these models~\cite{Lim_2017_CVPR_Workshops, Agustsson_2017_CVPR_Workshops, 2023LSDIRRN819, Nah_2019_CVPR_Workshops_REDS, 2019vimeoRN820}.

One of the most widely used datasets in this domain is \MIT, which provides RGB-D and hologram pairs designed for near-eye display scenarios~\cite{2021ShiRN222}.
Specifically, \MIT~offers 4,000 RGB-D image pairs and corresponding color holograms at resolutions of $192\times192$ and $384\times384$, with pixel pitches of 16 and 8~$\mu$m, respectively.
This dataset has served as a foundation for various ML-CGH studies, including hologram generation~\cite{2024FangRN781} and super-resolution~\cite{2022JeeRN500,2024NoRN499,2024LeeRN501}.
However, \MIT~has several limitations, most notably its narrow depth range of just 6~mm and its specific optical configuration, which is optimized for the Holoeye PLUTO device (8~$\mu$m pixel pitch and $1920\times1080$ Full HD resolution).

Other efforts have introduced private datasets to train their own ML-CGH models.
In particular, layer-based CGH methods using RGB-D inputs are widely adopted for dataset generation~\cite{2022yangRN695, 2024YanRN770, 2025YanRN791, 2024FangRN781, 2025LiaoRN783}, as they preserve the intensity of color images and facilitate direct comparison between holograms and rendered images at discrete depth planes.
These methods typically render RGB images and depth maps from multiple 3D objects and generate holograms via plane wave propagation using the angular spectrum method (ASM)~\cite{2022yangRN695, 2024YanRN770, 2025YanRN791}.
Alternatively, some approaches employ neural networks for depth estimation~\cite{depthanything}, allowing the creation of 3D hologram datasets from real-world RGB images without requiring explicit rendering of image-depth pairs~\cite{2024FangRN781, 2025LiaoRN783}.

Although these private datasets have shown strong performance in generating high-resolution holograms using ML, they are challenging for third-party researchers to reproduce.
Without access to the same datasets and hyperparameter configurations, training ML models often results in significantly different outcomes.
Additionally, many existing 3D holograms are composed of only a few depth layers and have limited depth coverage, restricting their ability to support immersive three-dimensional visualization.

These limitations underscore a critical gap: there is currently no publicly available dataset that offers both high resolution and a large depth range for training ML-based 3D CGH systems.
This lack of accessible, high-quality datasets hinders the development, benchmarking, and reproducibility of general-purpose ML-CGH models across the research community.

To address this gap, we introduce the \datasetname~dataset, a large-depth-range, high-resolution, and publicly available hologram dataset tailored for machine learning-based CGH.
It consists of 6,000 RGB-D and hologram pairs rendered at multiple resolutions (from $256\times256$ to $2048\times2048$), spanning a depth range of up to 80 mm, which is substantially larger than that of previous datasets.
The dataset is generated using a novel amplitude-projection-enhanced layer-based method, which improves reconstruction quality over existing techniques (Sec.~\ref{sec:AP-LBM}).
In addition to the dataset, we provide benchmarking experiments, including PSNR/SSIM comparisons, ML-based hologram generation, and super-resolution tasks.
Our contributions aim to catalyze progress in 3D ML-CGH by providing a robust and reproducible foundation for future research.

\section{Dataset Generation Pipeline}\label{sec:pipeline}




\begin{figure}
    \centering
    \includegraphics[width=1.0\linewidth]{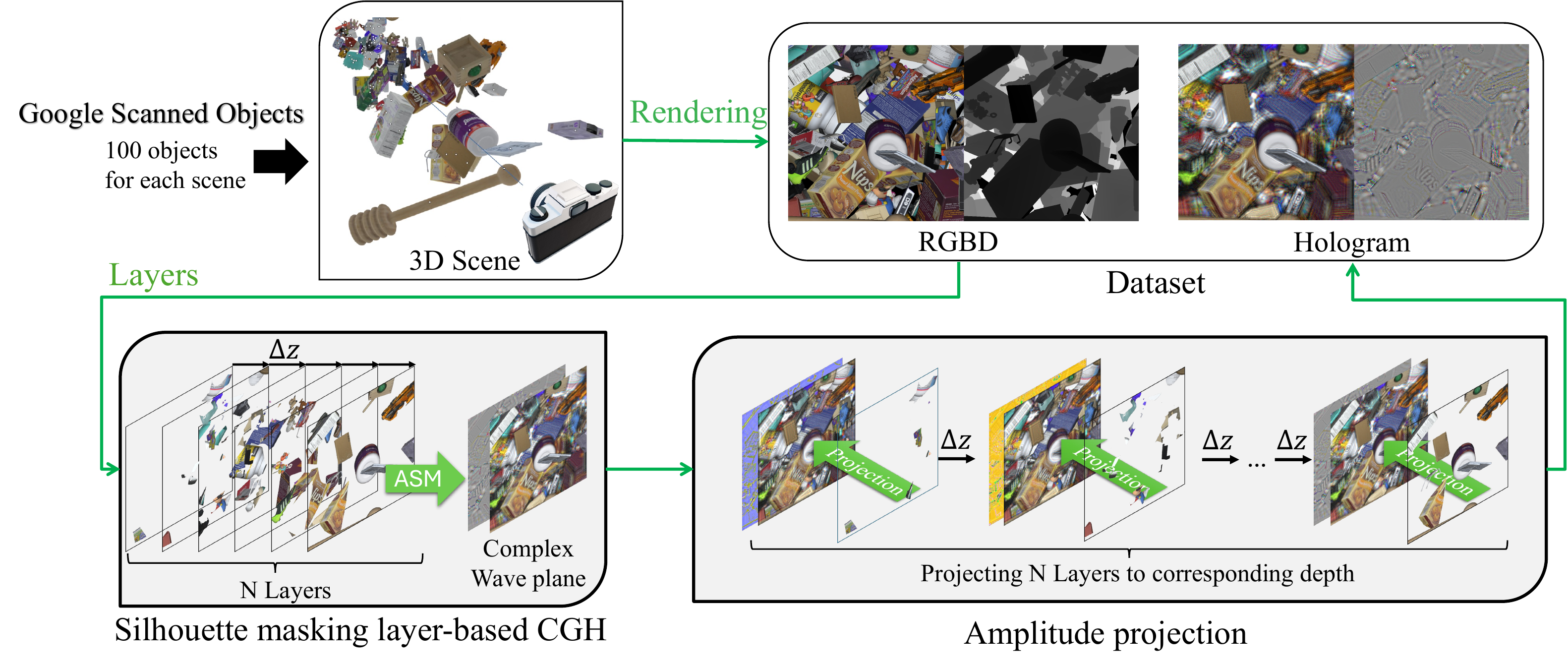}
    \caption{Overview of our dataset generation pipeline}
    \label{fig:overview}
\end{figure}
\begin{figure}
    \centering
    \subfigure[RGB]{
    \includegraphics[width=0.23\linewidth]{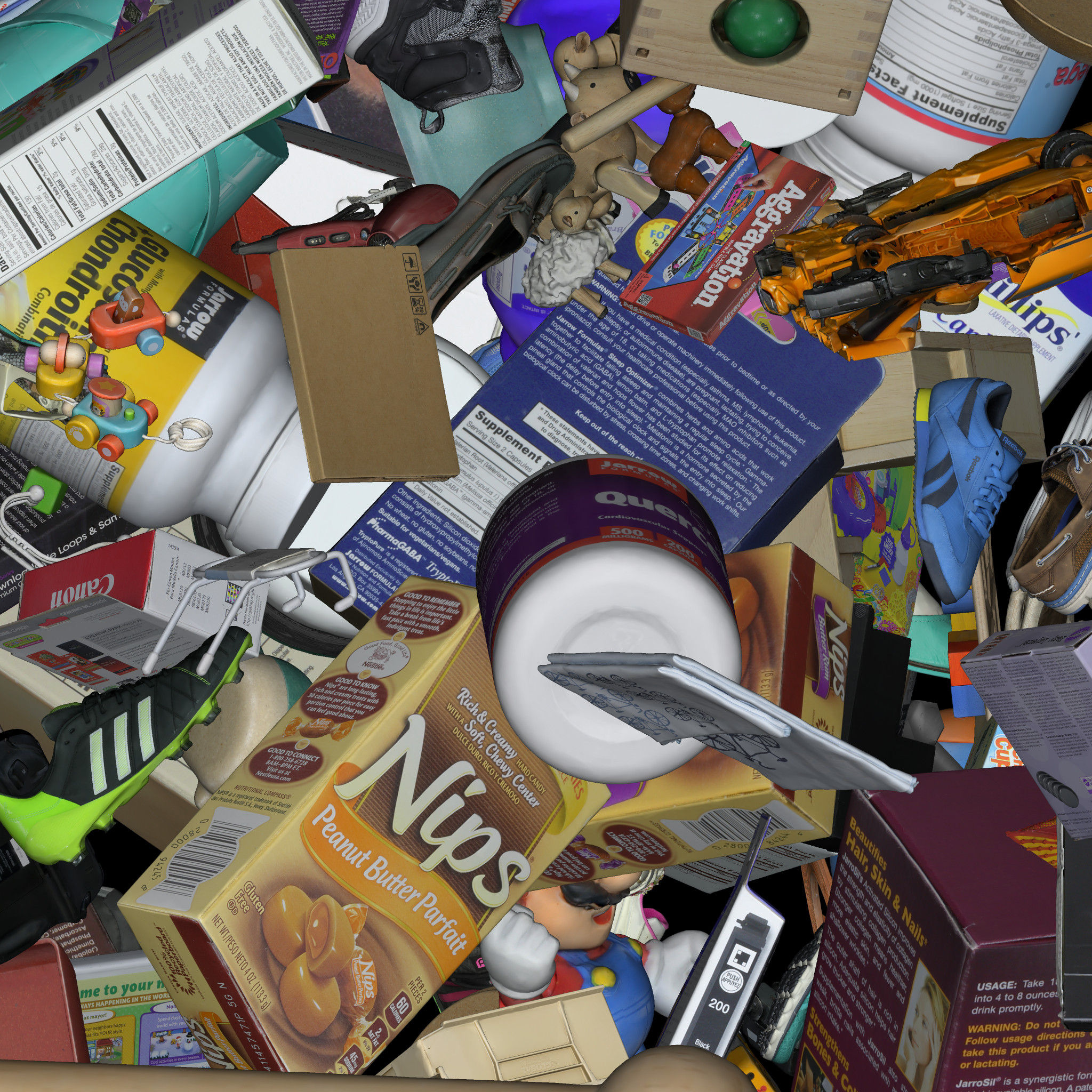}
    \label{fig:layerbased CGH comparison-rgb}
    }
    \subfigure[Depth map]{
    \includegraphics[width=0.23\linewidth]{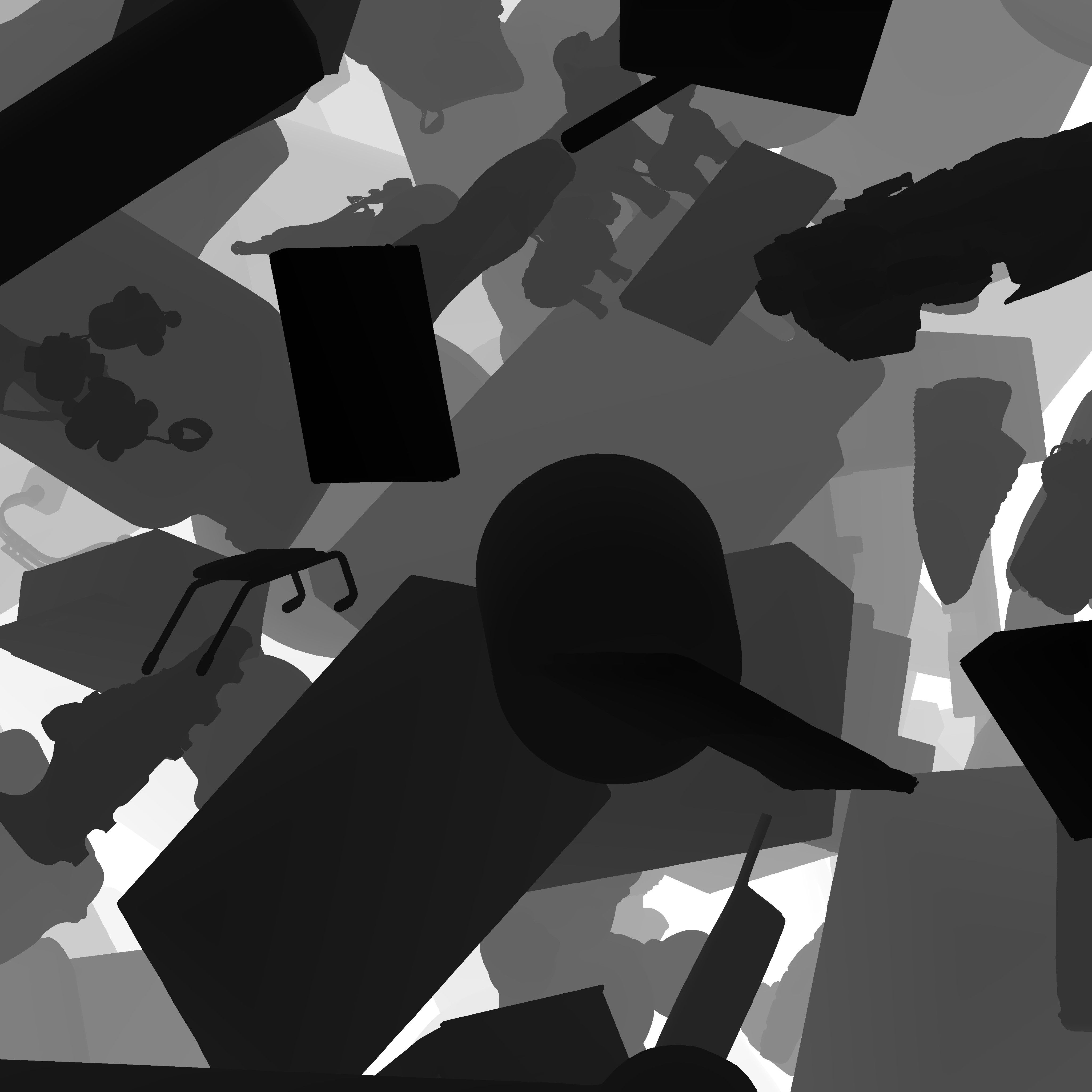}    
    \label{fig:layerbased CGH comparison-SM-LBM}
    }
    \subfigure[Amplitude]{
    \includegraphics[width=0.23\linewidth]{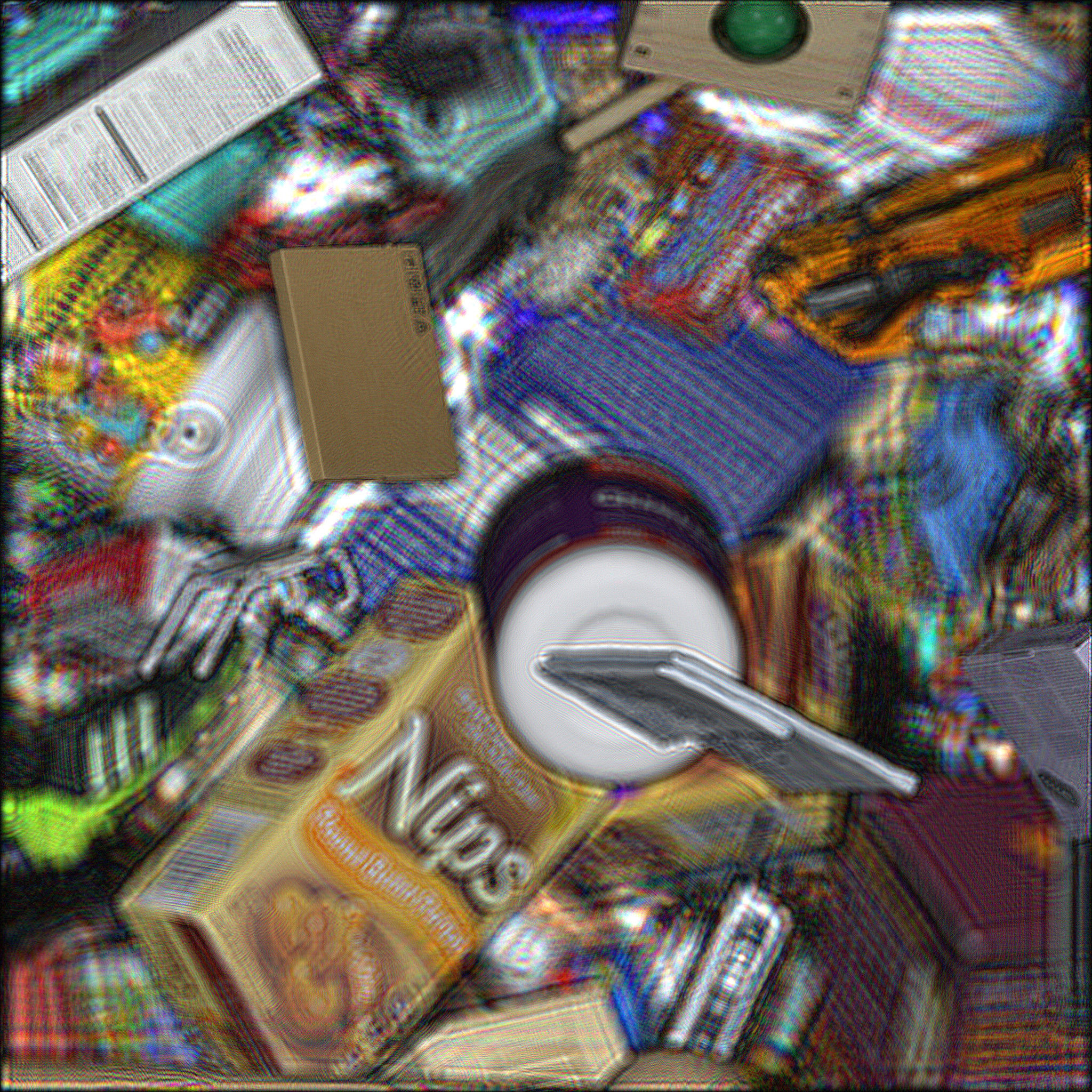}    
    \label{fig:layerbased CGH comparison-Adv}
    }
    \subfigure[Phase]{
    \includegraphics[width=0.23\linewidth]{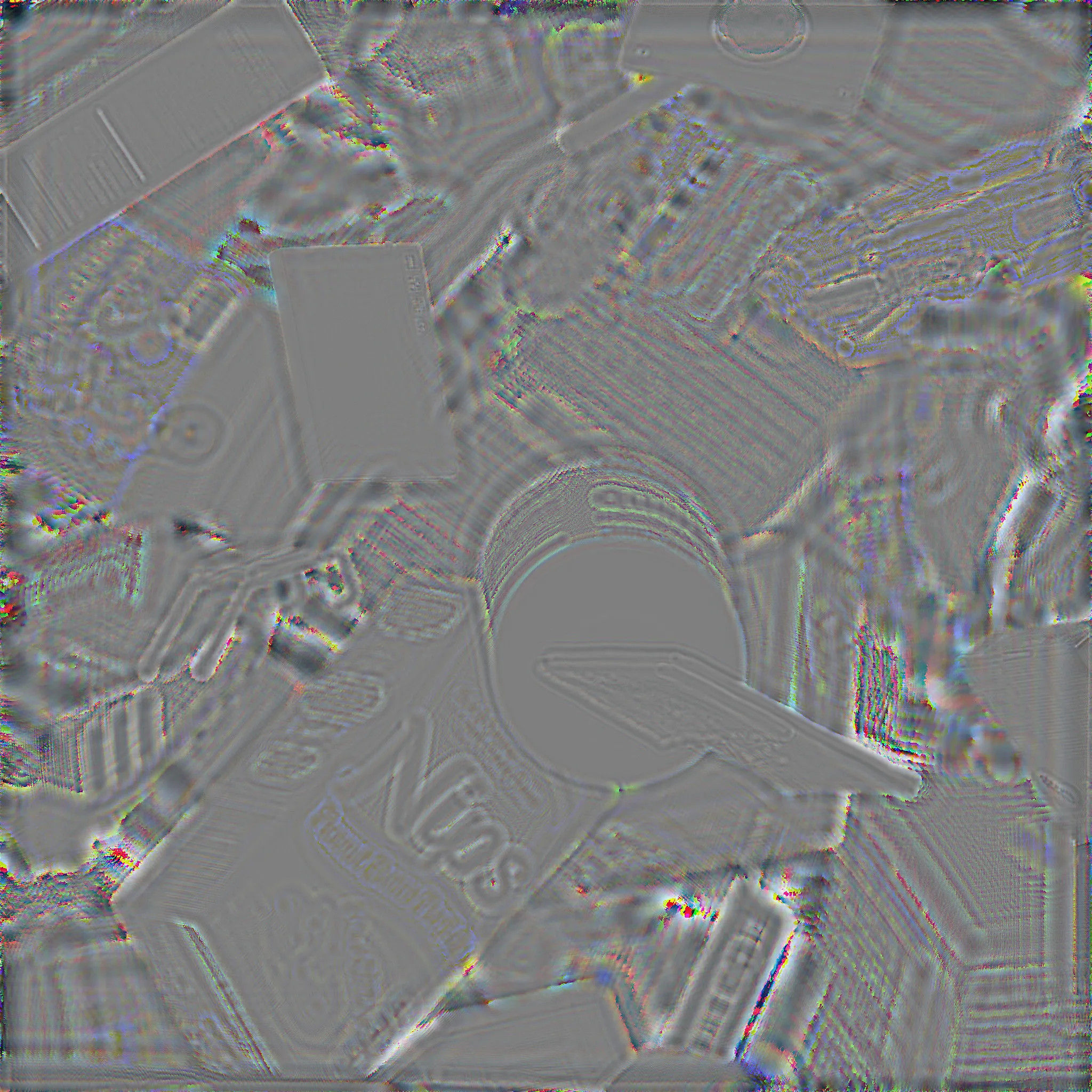}    
    \label{fig:layerbased CGH comparison-projection}
    }
    \caption{The configuration of \datasetname consists of (a) RGB and (b) depth map rendered from OptiX, and (c) amplitude and (d) phase of a hologram generated from AP-LBM.
    }
    \label{fig:data_sample}
\end{figure}

Each sample in the \datasetname~dataset consists of an RGB-D image and its corresponding hologram at multiple resolutions (Fig.~\ref{fig:data_sample}).
Figure~\ref{fig:overview} illustrates the overall pipeline used to generate the \datasetname~dataset.

The scenes are composed of 100 objects selected from the Google Scanned Objects (GSO) dataset~\cite{2022googleRN497}, a high-quality collection of 3D-scanned models designed for 3D machine learning tasks.
To ensure scene diversity, each object is randomly transformed and positioned within the camera's field of view.

\begin{figure}
    \centering
    \subfigure[Objects' XY-position]{
    \includegraphics[width=0.25\linewidth]{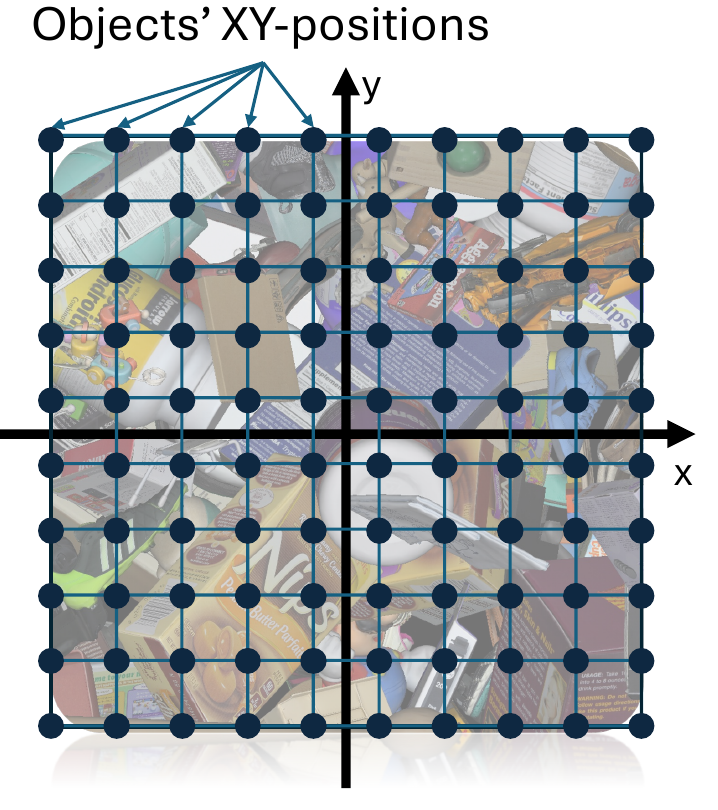}
    \label{fig:xy}
    }
    \subfigure[Objects' Z-position]{
    \includegraphics[width=0.32\linewidth]{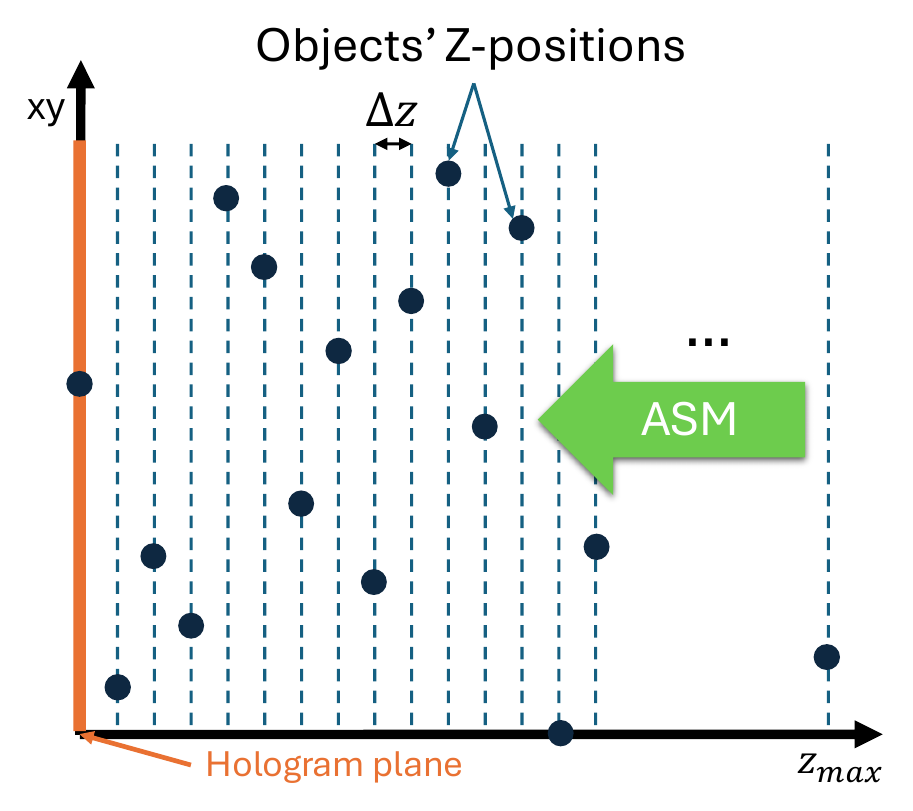}    
    \label{fig:z}
    }
    \caption{
    Object placement in the 3D scene.
    The figure shows the center positions of the placed objects in the XY-plane and their corresponding depths along the Z-axis.
    }
    \label{fig:position}
\end{figure}

First, each object is resized to a scale between 20\% and 30\% of the hologram width.
Next, random rotations are applied using Euler angles sampled from the range $[-2\pi, 2\pi)$ in yaw, pitch, and roll.
The transformed objects are then placed on a two-dimensional uniform grid in the camera plane, as shown in Fig.~\ref{fig:position}.
This grid ensures even distribution along the $x$- and $y$-axes.
For the $z$-axis, the position of each object is independently sampled from the interval $z_\text{object} \in (-z_\text{max}, z_\text{min}]$, where $z_\text{min} = 0$ and $z_\text{max}$ is the maximum effective propagation distance of the angular spectrum method (ASM), defined as:

\begin{equation}
z_\text{max} \leq  N \Delta x \sqrt{4(\Delta x / \lambda)^2 - 1}
\end{equation}

Here, $N$ is the number of pixels along the width or height of the hologram, $\Delta x$ is the pixel pitch, and $\lambda$ is the wavelength of light.

Scene rendering was performed using OptiX 8.0 with orthographic projection, producing single-precision RGB intensities and depth maps.
The wavelength $\lambda$ corresponds to the red channel (e.g., 638 nm), which determines the shortest valid $z_\text{max}$ used in the rendering configuration.

From the rendered RGB-D images, we generated holograms using a silhouette-masking layer-based hologram generation method~\cite{2017ZhangRN798, 2025LiaoRN783, 2025duanRN800}, enhanced with amplitude projection (Sec.~\ref{sec:AP-LBM}).
These holograms were synthesized for large-depth-range scenes, ensuring all layers remain within the theoretical propagation limits of numerical diffraction~\cite{2015ZhaoRN87}.

\section{Large-Depth-Range Hologram Generation}
\label{sec:AP-LBM}

In this section, we explain how high-quality holograms with large depth ranges are generated using a novel amplitude-projection-enhanced layer-based method (AP-LBM).
Specifically, AP-LBM combines two components: an advanced silhouette-masking layer-based CGH method (Sec.~\ref{subsec:SM-LBM}) and an amplitude projection technique (Sec.~\ref{subsec:AP}).

\subsection{Advanced silhouette-masking layer-based CGH}
\label{subsec:SM-LBM}

The first step of AP-LBM is the silhouette-masking layer-based CGH method.
This approach samples light wavefronts from multiple two-dimensional planes along the $z$-axis and numerically propagates them to the hologram plane.
To ensure that only the appropriate wavefronts contribute to the final hologram, the silhouette masking method~\cite{2009MatsushimaRN13, 2017ZhangRN798} is applied to filter out irrelevant wave planes.


The original silhouette-masking layer-based method (SM-LBM) is formulated as:
\begin{equation}
        U_{i+1}(x,y) = I(x,y)\circ M_{i}(x,y)\exp{(-\sqrt{-1}z_i)} + prop.(U_i(x,y),-\Delta z)\circ \sum_{k=0}^{i-1}M_{i}(x,y)
    \label{eq:original_SM-LBM}
\end{equation}
Here, $U_i(x, y)$ is the $i$\textsuperscript{th} complex wavefront, and $I(x, y)$ is the RGB image rendered from the target 3D scene.
$M_i(x, y)$ is the $i$\textsuperscript{th} binary masking plane, where a value of 1 indicates that an object exists at depth $z_i$, and 0 otherwise.
The operator $\circ$ denotes the Hadamard (element-wise) product.
$z_i$ is the physical distance between the hologram plane and the $i$\textsuperscript{th} layer, and $\Delta z$ is the inter-layer spacing.
$U_{n-1}(x, y)$ represents the hologram plane after propagating through $n$ layers, and the farthest wavefront $U_0(x, y)$ is initialized as:
\begin{equation}
    U_{0}(x,y) = I(x,y)\circ M_{0}(x,y)\exp{(-\sqrt{-1}z_0)}
\end{equation}
The operator \textit{prop.} denotes numerical propagation.
We use the angular spectrum method (ASM)~\cite{2005GoodmanRN277} for plane-to-plane propagation, defined as:
\begin{equation}
    prop.(U(x,y), z) = IDFT \left[DFT\{U(x,y)\}
    \cdot\exp{(2\sqrt{-1}\pi z\sqrt{1/{\lambda}^2- f_x^2-f_y^2})} \right]
\end{equation}
Here, $f_x$ and $f_y$ are the spatial frequencies in the Fourier domain, $z$ is the propagation distance, and $\lambda$ is the wavelength of light.
We apply FFT-shift and zero-padding during the DFT (discrete Fourier transform) and IDFT (inverse discrete Fourier transform) operations to reduce aliasing effects and improve accuracy.


One limitation of SM-LBM is the appearance of contour line aliasing between focal planes and visible shadow artifacts near object boundaries in the reconstructed holograms.
To mitigate these issues, we incorporate advanced layer-based CGH techniques~\cite{2025LiaoRN783, 2025duanRN800}, which improve hologram quality, especially for large-depth-range scenes.

The improved generation method is given by:
\begin{equation}
    U_{i+1}(x,y) = I(x,y)\circ M_{i,2}(x,y)\exp{(-\sqrt{-1}z_i)}
    + prop.(U_i(x,y),-\Delta z)\circ M_{i,2}^{-1}
    \label{eq:hologram_generation}
\end{equation}

The masking matrix $M_{i,k}(x, y)$ for the $i$\textsuperscript{th} layer is defined as:
\begin{equation}
M_{i,k}(x,y) = 
\begin{cases}
1, &\text{if } \max(D) - i\Delta z \ge D(x,y) > \max(D) - (i + k)\Delta z,
     \\
0, & \text{otherwise}.
\end{cases}
\label{eq:bi-directionalMasking}
\end{equation}

Its inverse is given by:
\begin{equation}
    M_{i,k}^{-1}(x,y)= 
    \begin{cases}
    0,& if ~M_{i,k}(x,y) = 1,  \\
    1, &\text{otherwise}.
    \end{cases}
\end{equation}

Here, $D(x,y)$ is the depth map corresponding to the rendered image $I(x,y)$, and $\max(D)$ represents the maximum depth value in the scene.
The parameter $k$ defines the number of layers used for masking; we set $k=2$ to suppress contour lines between layers~\cite{2025LiaoRN783}.
The farthest plane, $U_0(x,y)$, is given by 
\begin{equation}
    U_{0}(x,y) = I(x,y)\circ M_{0,2}(x,y)\exp{(-\sqrt{-1}z_0)}
\end{equation}

In contrast to the original SM-LBM, which masks only backward layers (e.g., $D(x,y) > \max(D) - (i + k)\Delta z$), the bidirectional masking approach in Eq.~\ref{eq:hologram_generation} considers both forward and backward directions relative to the current focal layer~\cite{2025duanRN800}.
This leads to improved visual quality in the reconstructed images and effectively reduces exceptional shadow artifacts caused by silhouette masking.


\subsection{Amplitude projection}
\label{subsec:AP}

When the depth range of a hologram scene becomes very large, defocus blur from the front planes can obscure details in the background, significantly degrading the quality of the reconstructed image.
To address this issue, we propose an amplitude projection method inspired by alternate projection techniques~\cite{2024nonconvexRN799}, which projects the image onto the amplitude component of the hologram wavefield at its corresponding focal plane.

The amplitude projection process is defined as:
\begin{equation}
\begin{split}
     U'_{i+1}(x,y)= 
    I(x,y)\circ M_{i,1}(x,y) 
    \circ \exp\left[\sqrt{-1} \cdot |prop.(U_{i}^{'}(x,y),-\Delta z)|_\Phi\right] \\
    + M_{i,1}^{-1}(x,y)\circ prop.(U_{i}^{'}(x,y),-\Delta z )
\end{split}
        \label{eq:projection}
\end{equation}
\begin{equation}
\begin{split}
    U'_{0}(x,y)= 
    I(x,y)\circ M_{0,1}(x,y) 
    \circ \exp\left[\sqrt{-1} \cdot |prop.(U_{n-1} (x,y),\max(D))|_\Phi\right]\\
    + M_{0,1}^{-1}(x,y)\circ prop.(U_{n-1}(x,y),\max(D))
        \end{split}
        \label{eq:projection_0}
\end{equation}
Here, $U'_i(x,y)$ is a refined hologram plane obtained by applying amplitude projection at the $i$\textsuperscript{th} layer, and $|\cdot|_\Phi$ represents the phase component of the complex value.
The first term in Eq.~\ref{eq:projection} replaces the amplitude components of $U'_i(x,y)$ within the masked region defined by $M_{i,1}(x,y)$ while preserving their phase.
The second term retains the unchanged values at the remaining regions of the $i$\textsuperscript{th} plane.

The process begins with $U_{n-1}(x, y)$, the hologram generated by the advanced layer-based method (Eq.~\ref{eq:hologram_generation}).
This hologram is first propagated to the farthest focal plane (i.e., $U_0$) using numerical diffraction.
At that depth, the amplitude is replaced with the image content corresponding to the same depth.
The the focal plane is then propagated by $-\Delta z$ to the next depth layer ($U_1$), where the amplitude component is again updated using the masking matrix $M_{1,1}(x, y)$.
This iterative process continues until the refined hologram plane $U'_{n-1}(x, y)$ is reached.

Experimental results presented in Sec.~\ref{sec:quality_eval} show that this amplitude projection method outperforms conventional layer-based CGH techniques.
It yields superior quantitative metrics and enhances reconstruction quality, particularly at deeper focal planes.



\section{Hologram Quality Evaluation}
\label{sec:quality_eval}

In this section, we evaluate the hologram quality produced by our proposed method, AP-LBM, in comparison with baseline layer-based CGH approaches.
Specifically, we consider the following three methods:
\begin{itemize}
    \item \textbf{SM-LBM}: the original silhouette-masking layer-based method (Eq.~\ref{eq:original_SM-LBM})
    \item \textbf{ADV-LBM}: the advanced silhouette-masking method incorporating bidirectional masking (Sec.~\ref{subsec:SM-LBM}, Eq.~\ref{eq:hologram_generation})
    \item \textbf{AP-LBM}: the full proposed method, which further integrates amplitude projection (Sec.~\ref{subsec:AP}, Eq.~\ref{eq:projection})
\end{itemize}

We perform both numerical and optical reconstructions to qualitatively assess visual fidelity.
Additionally, we conduct quantitative evaluations using PSNR and SSIM to compare reconstruction accuracy across different methods.


\subsection{Dataset configuration}
\label{subsec:dataset}

\begin{table}[htbp]
\centering
\caption{Physical extents of the hologram volume at different resolutions}
\begin{tabular}{c|c|c|c}
\hline
Resolution & Width (mm) & Height (mm) & Depth Range (mm) \\
\hline
$256 \times 256$   & 0.9216 & 0.9216 & 10.1668        \\
$512 \times 512$   & 1.8432 & 1.8432 & 20.3336  \\
$1024 \times 1024$ & 3.6864 & 3.6864 & 40.6672   \\
$2048 \times 2048$ & 7.3728 & 7.3728 & 81.3344	  \\
\hline
\end{tabular}
\label{tab:volume_dimensions}
\end{table}

The \datasetname~dataset includes RGB-D and complex hologram pairs at four spatial resolutions: $256^2$, $512^2$, $1024^2$, and $2048^2$. All samples are generated using a fixed pixel pitch of 3.6~$\mu$m and wavelengths of 638~nm, 532~nm, and 450~nm, corresponding to red, green, and blue channels, respectively.
The maximum propagation distance $z_{\text{max}}$ for each resolution is computed using the angular spectrum method (ASM), as defined in Eq.~\ref{eq:original_SM-LBM}.
For the $1024^2$ case, the physical dimensions of the hologram volume are approximately $3.6864 \times 3.6864 \times 40.6672$~mm$^3$ (Table.~\ref{tab:volume_dimensions}).
Each hologram is generated from the RGB-D data using the proposed AP-LBM method (Eq.~\ref{eq:projection}) with 10,000 layers. Data are stored in 32-bit floating-point format: RGB, depth, and phase values are normalized to the [0, 1] range, while amplitude values vary based on scene content.

\subsection{Qualitative evaluation}
\label{sec:result_qualtitative}

To evaluate the quality of holograms generated by our proposed method (i.e., AP-LBM), we generated holograms with a resolution of $2048 \times 2048$, corresponding to a scene volume of $7.37 \times 7.37 \times 81.33~\text{mm}^3$ in width, height, and depth, respectively.
We conducted this evaluation using three different algorithms: SM-LBM, ADV-LBM, and AP-LBM.


\begin{figure}
    \centering
    \subfigure[SM-LBM (focus: front)]{
    \includegraphics[width=0.4\linewidth]{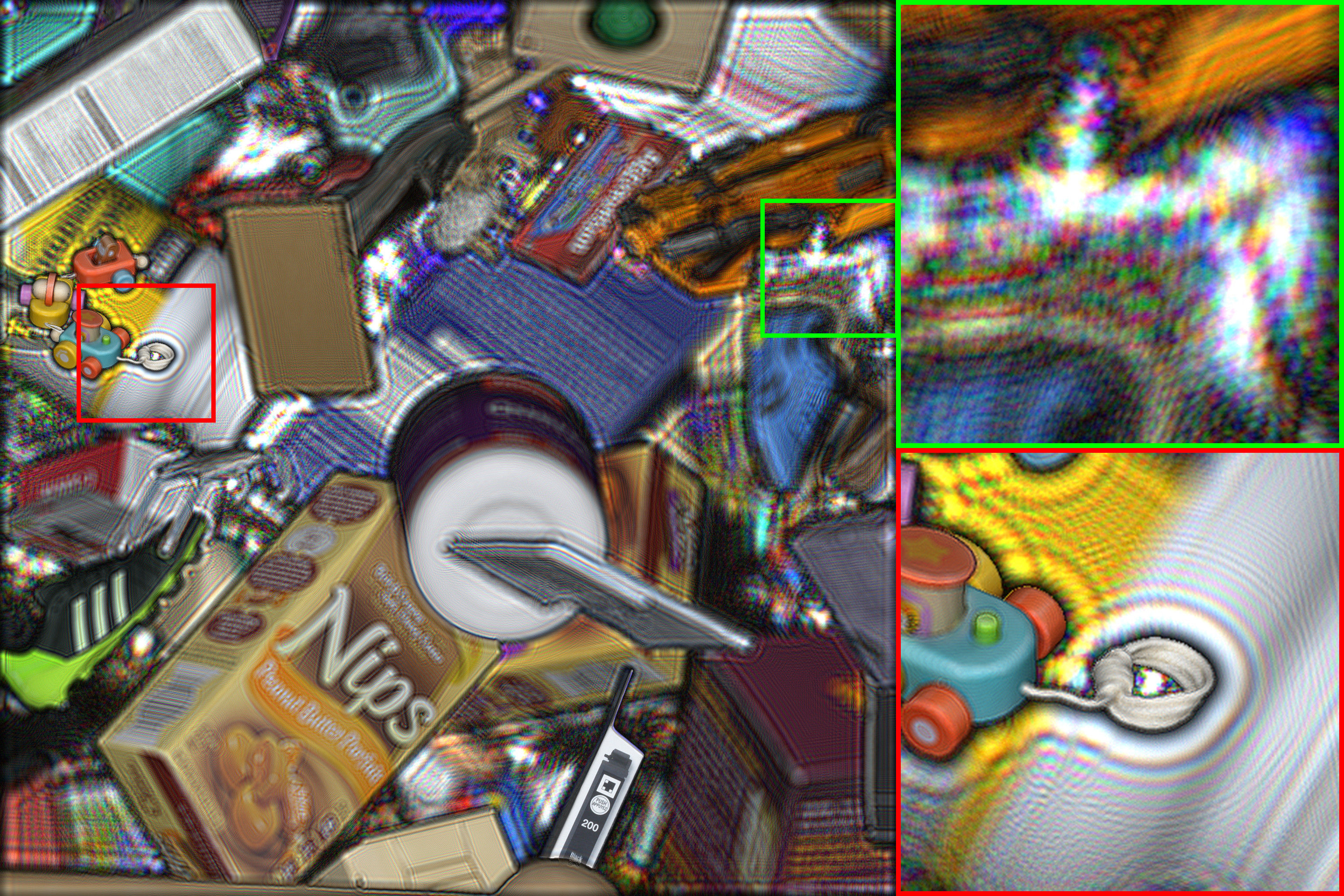}    
    \label{fig:layerbased CGH comparison-SM-LBM}
    }
    \subfigure[SM-LBM (focus: back)]{
    \includegraphics[width=0.4\linewidth]{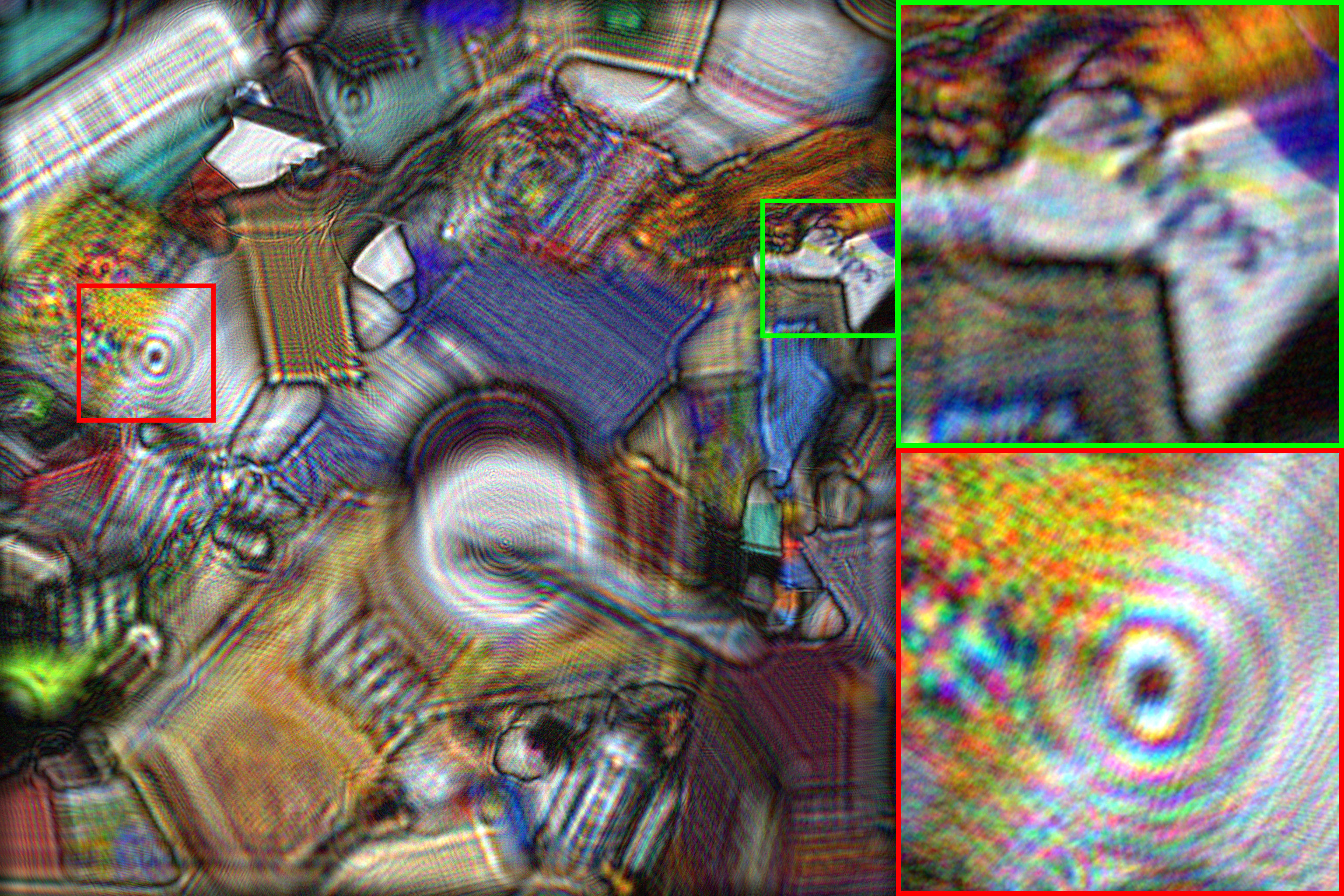}    
    \label{fig:layerbased CGH comparison-Adv}
    }
    \subfigure[ADV-LBM (focus: front)]{
    \includegraphics[width=0.4\linewidth]{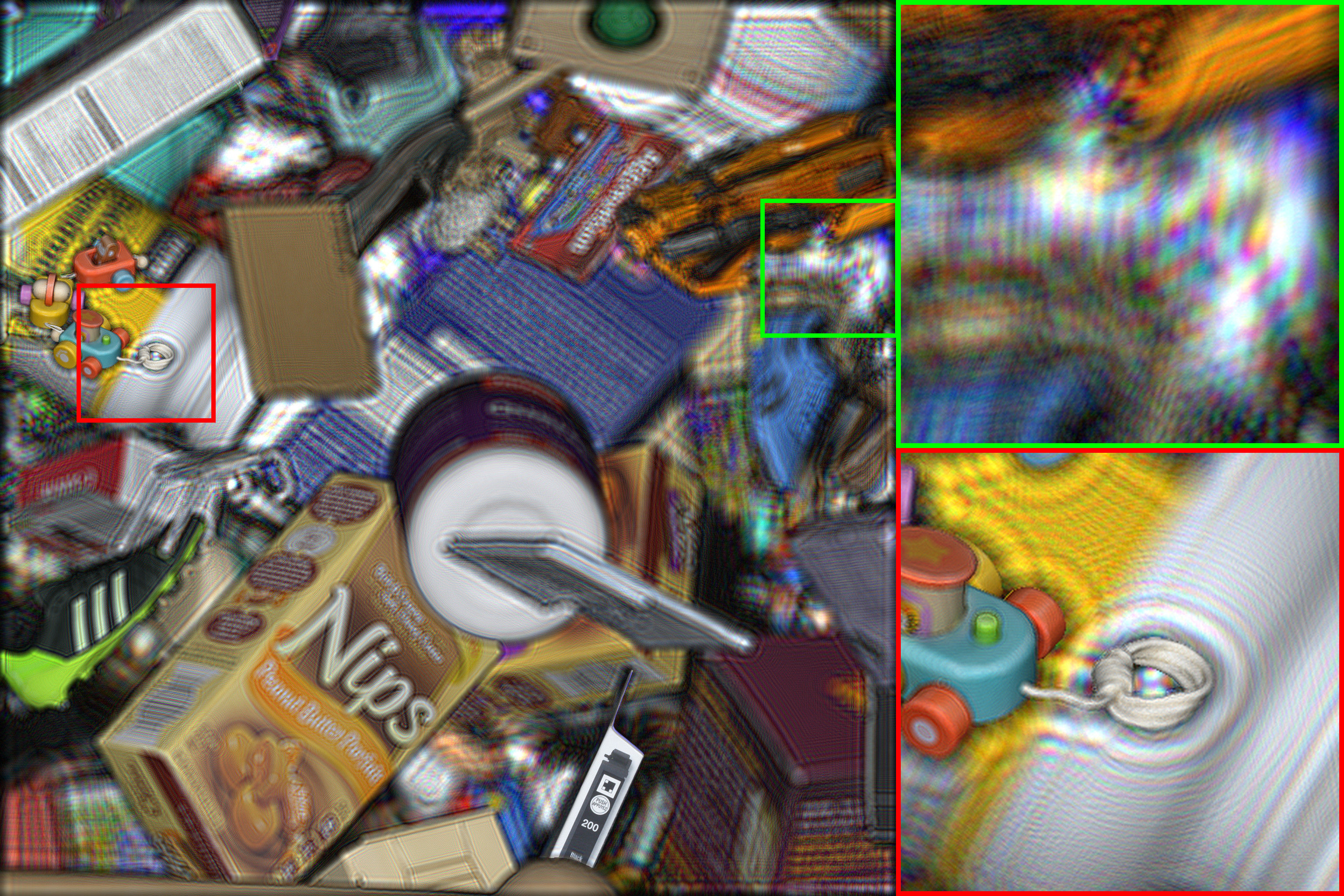}    
    \label{fig:layerbased CGH comparison-projection}
    }
    \subfigure[ADV-LBM (focus: back)]{
    \includegraphics[width=0.4\linewidth]{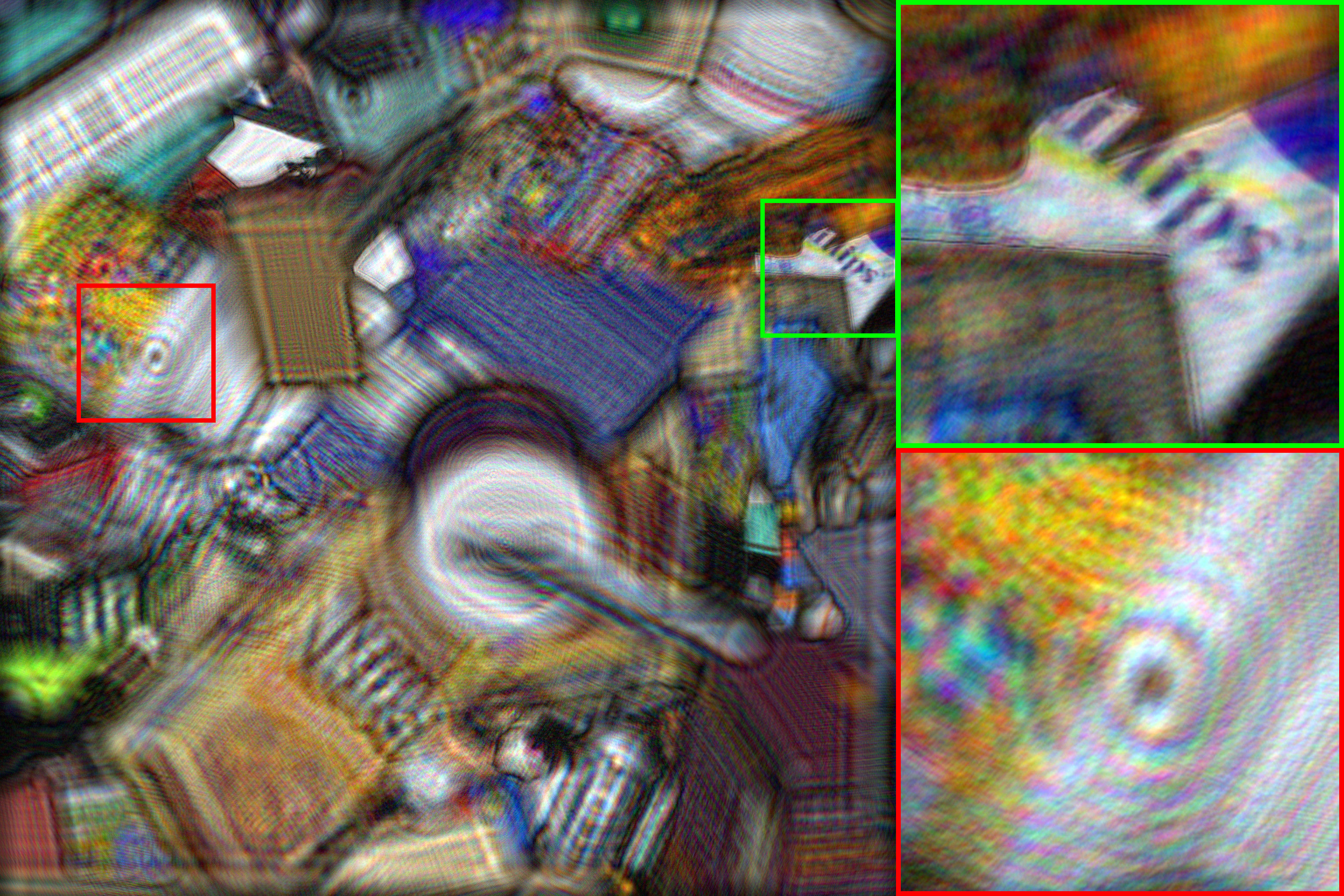}    
    \label{fig:layerbased CGH comparison-SM-LBM}
    }
    \subfigure[AP-LBM (focus: front)]{
    \includegraphics[width=0.4\linewidth]{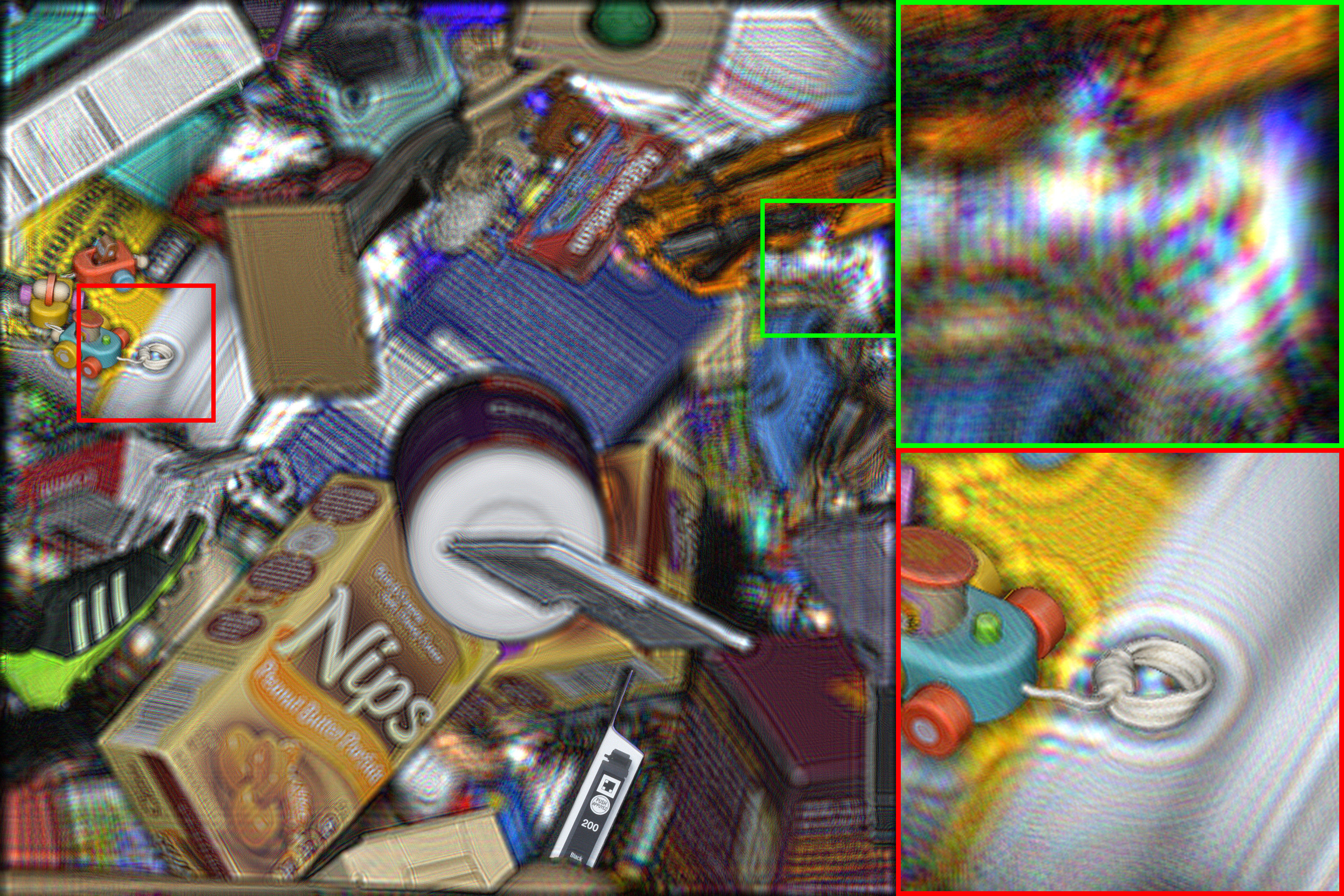}    
    \label{fig:layerbased CGH comparison-Adv}
    }
    \subfigure[AP-LBM (focus: back)]{
    \includegraphics[width=0.4\linewidth]{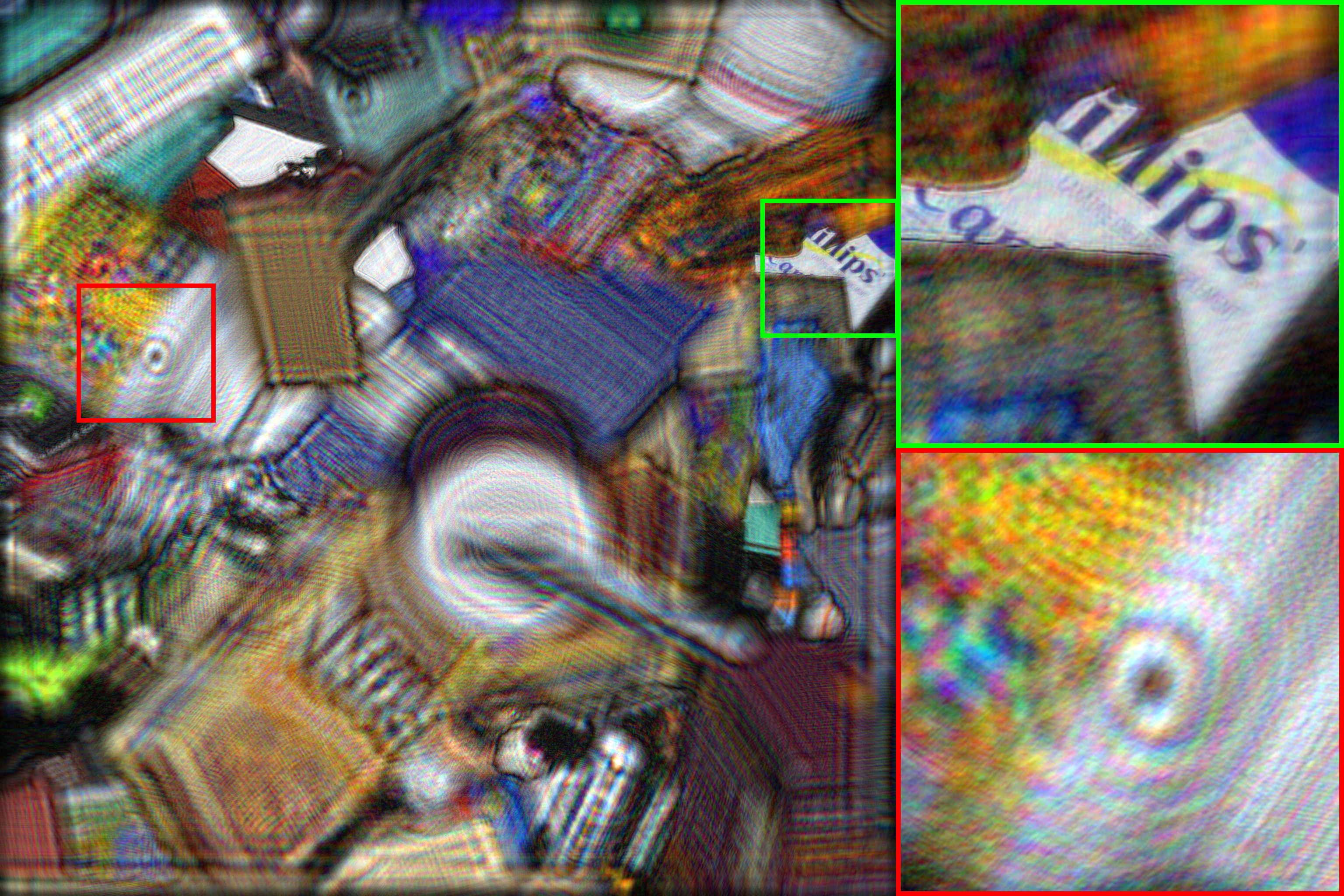}    
    \label{fig:layerbased CGH comparison-projection}
    }
    \caption{Numerical reconstructions of holograms generated using different methods, with focus set to front (left) and back (right) planes.
    }
    \label{fig:numerical_recon}
\end{figure}

Fig.~\ref{fig:numerical_recon} presents the numerical reconstructions of holograms generated using the three methods.
The SM-LBM method exhibits noticeable outline artifacts, with object contours remaining visible even in defocused regions.
In contrast, both ADV-LBM and AP-LBM significantly reduce these artifacts.
In the back-focused regions, SM-LBM and ADV-LBM display considerable blurring, even for in-focus details, whereas AP-LBM preserves sharper features, particularly in fine structures such as text.
We observed similar trends across multiple scenes, confirming the consistent advantage of AP-LBM in numerical reconstruction quality.

\begin{figure}
    \centering
    \includegraphics[width=0.8\linewidth]{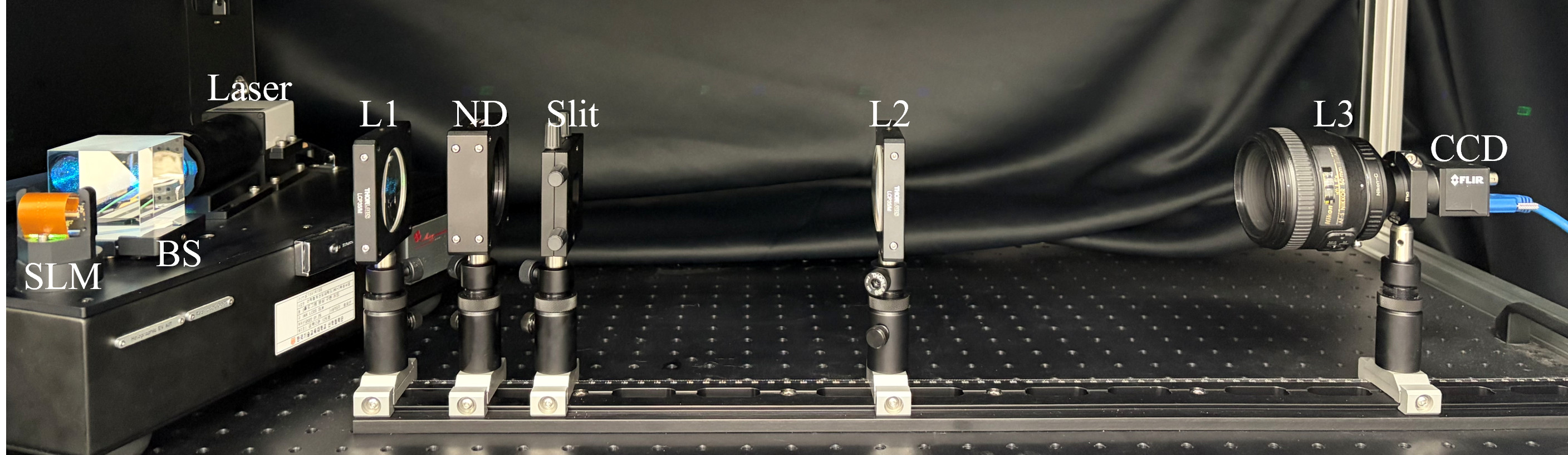}
    \caption{
    Optical setup used for hologram reconstruction. The system includes Lens 1 (focal length: 100 mm) and Lens 2 (focal length: 200 mm), an ND filter (optical density: 3.0), an adjustable slit (OWIS SP60), and a camera lens (Nikon 50 mm), paired with a FLIR Blackfly S camera.
    }
    \label{fig:optical_config}
\end{figure}
\begin{figure*}
    \centering
    \subfigure[SM-LBM (focus: front)]{
    \includegraphics[width=0.4\linewidth]{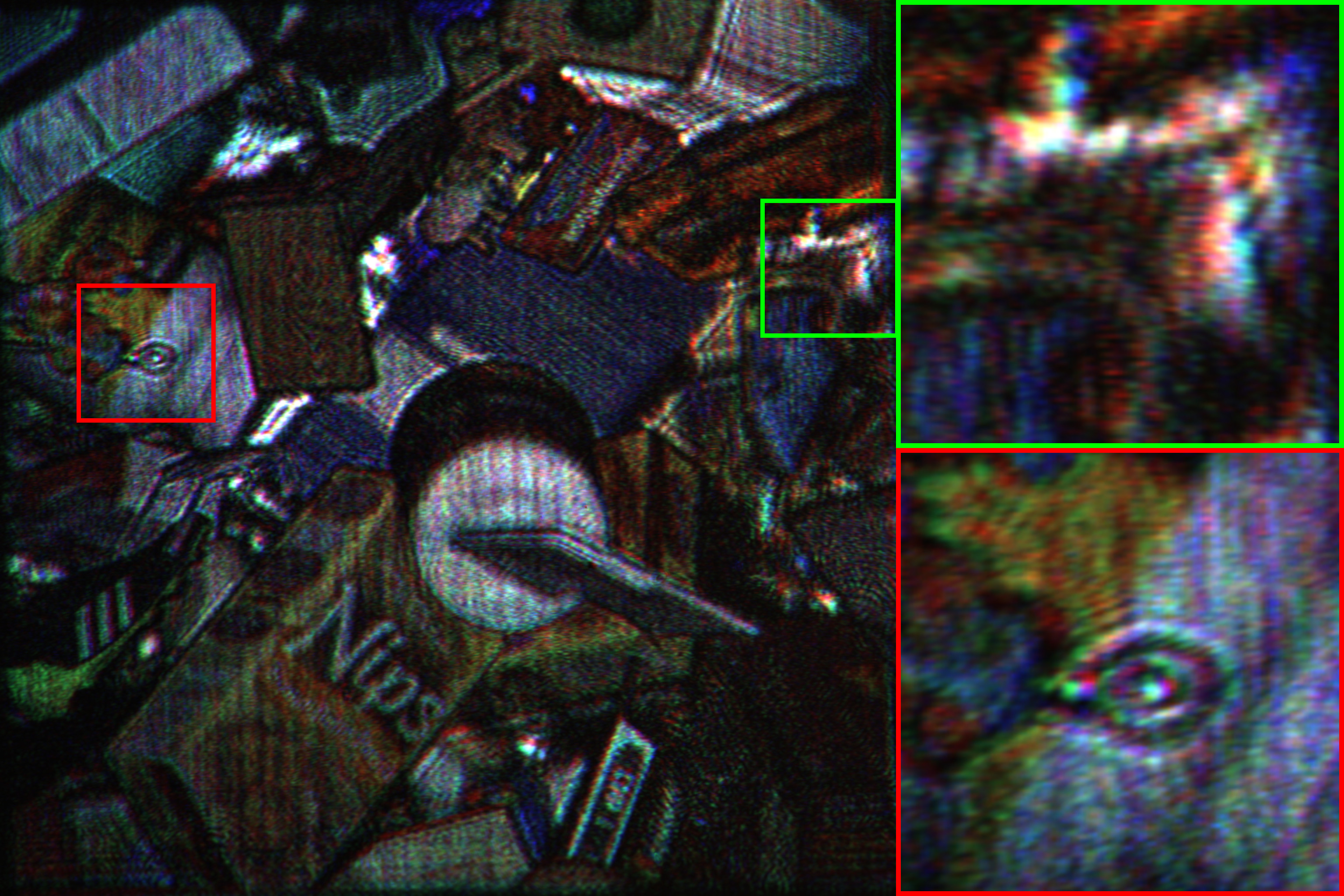}    
    \label{fig:layerbased CGH comparison-SM-LBM}
    }
    \subfigure[SM-LBM (focus: back)]{
    \includegraphics[width=0.4\linewidth]{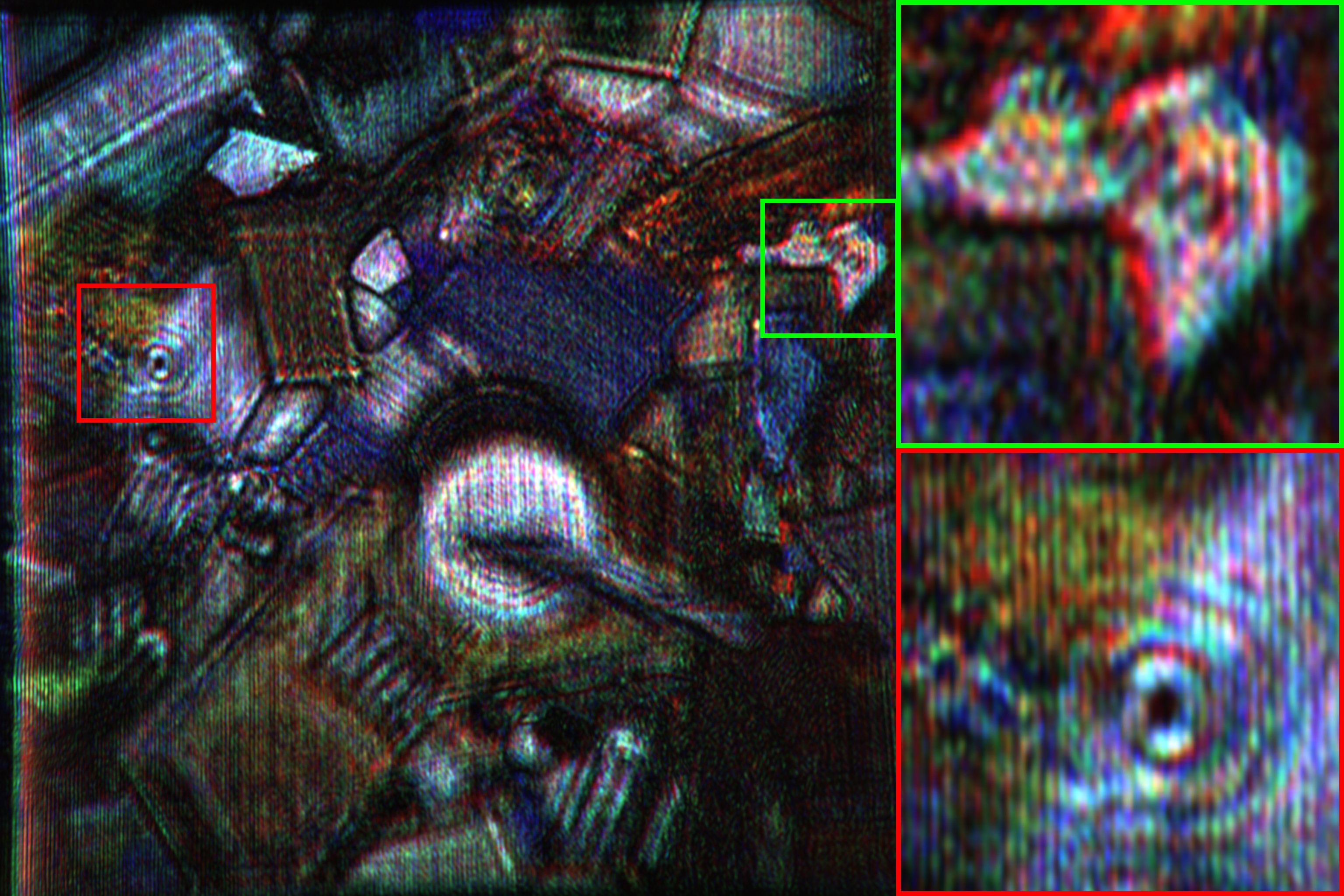}
    \label{fig:layerbased CGH comparison-Adv}
    }
    \subfigure[ADV-LBM (focus: front)]{
    \includegraphics[width=0.4\linewidth]{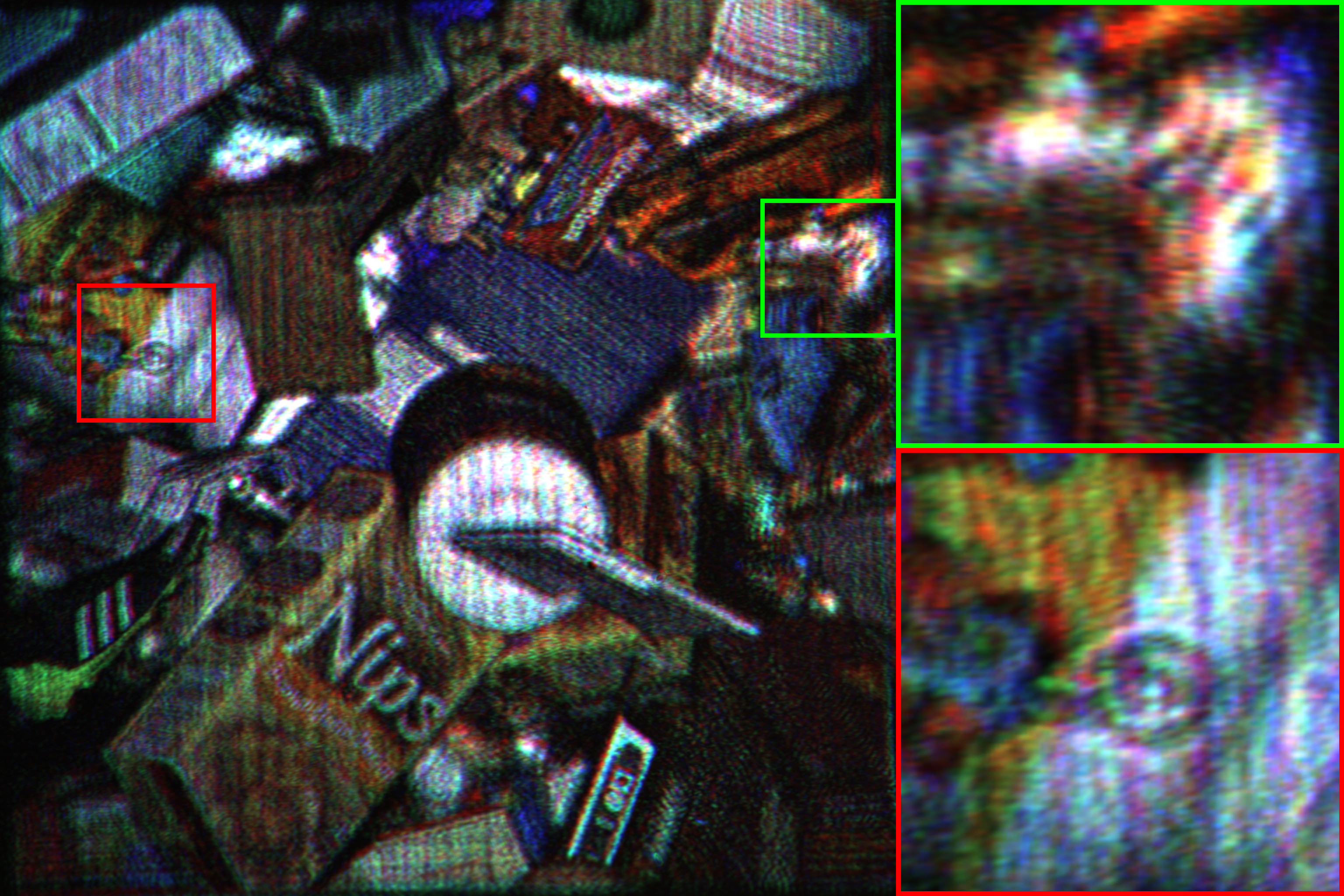}    
    \label{fig:layerbased CGH comparison-projection}
    }
    \subfigure[ADV-LBM (focus: back)]{
    \includegraphics[width=0.4\linewidth]{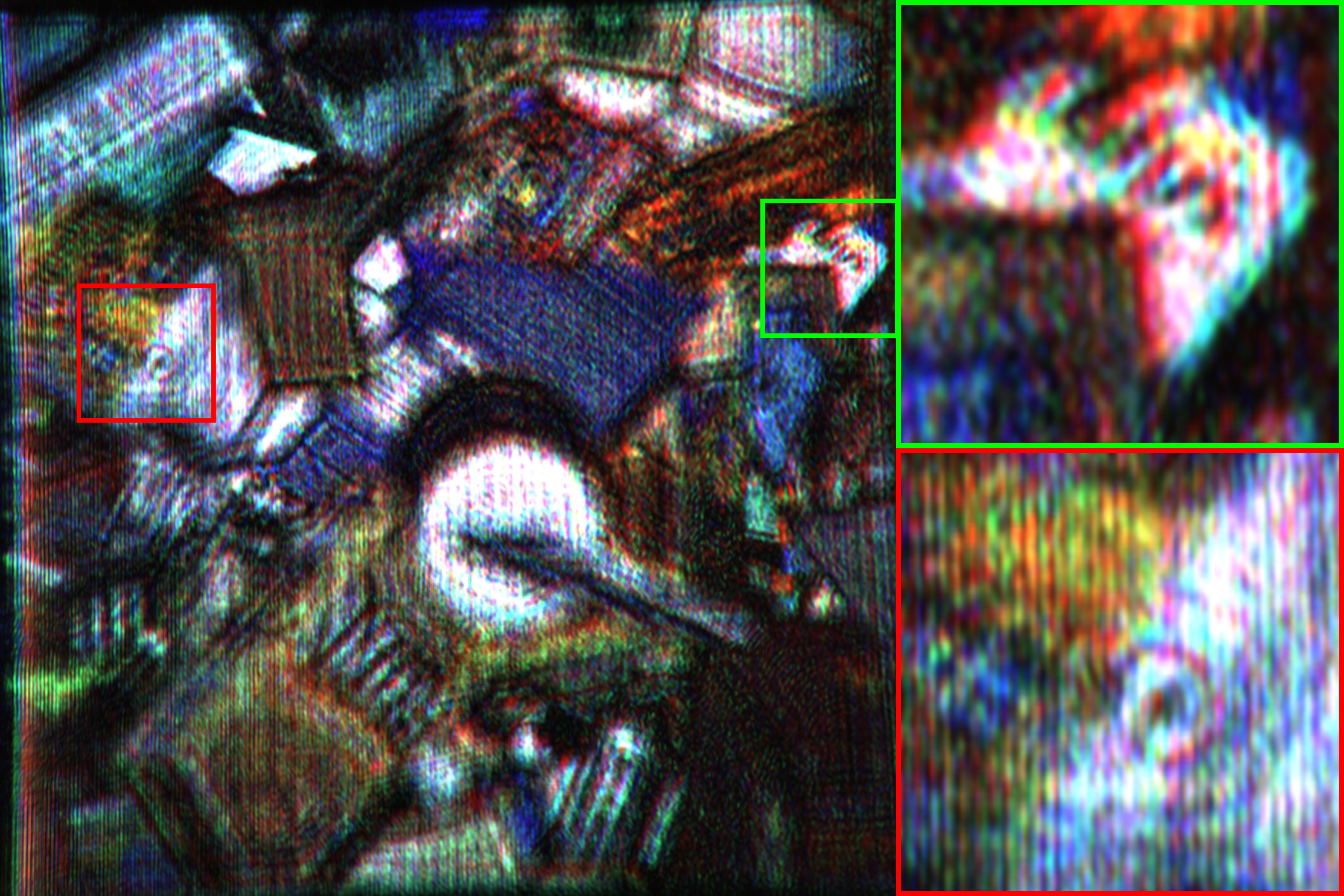}    
    \label{fig:layerbased CGH comparison-SM-LBM}
    }
    \subfigure[AP-LBM (focus: front)]{
    \includegraphics[width=0.4\linewidth]{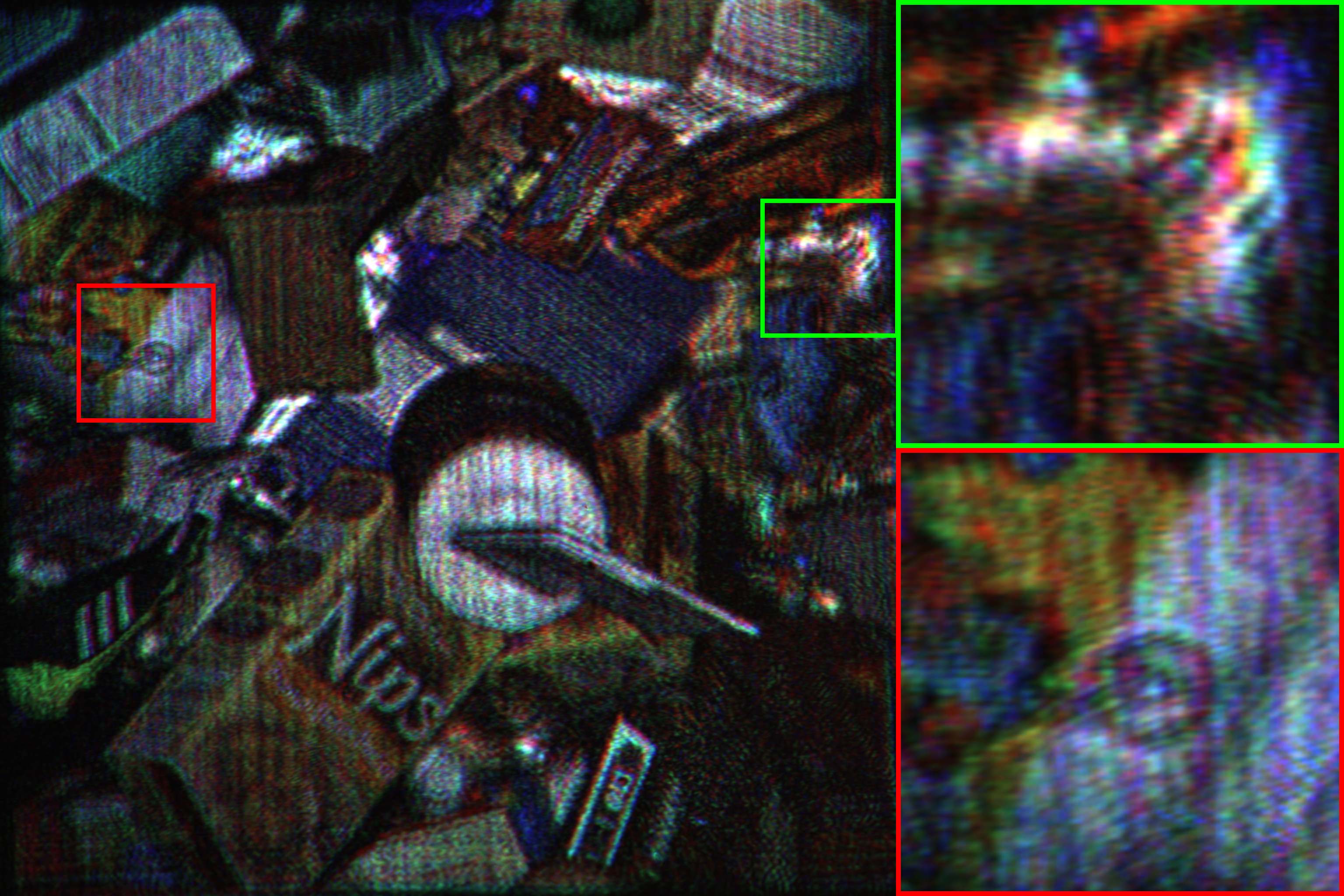}    
    \label{fig:layerbased CGH comparison-Adv}
    }
    \subfigure[AP-LBM (focus: back)]{
    \includegraphics[width=0.4\linewidth]{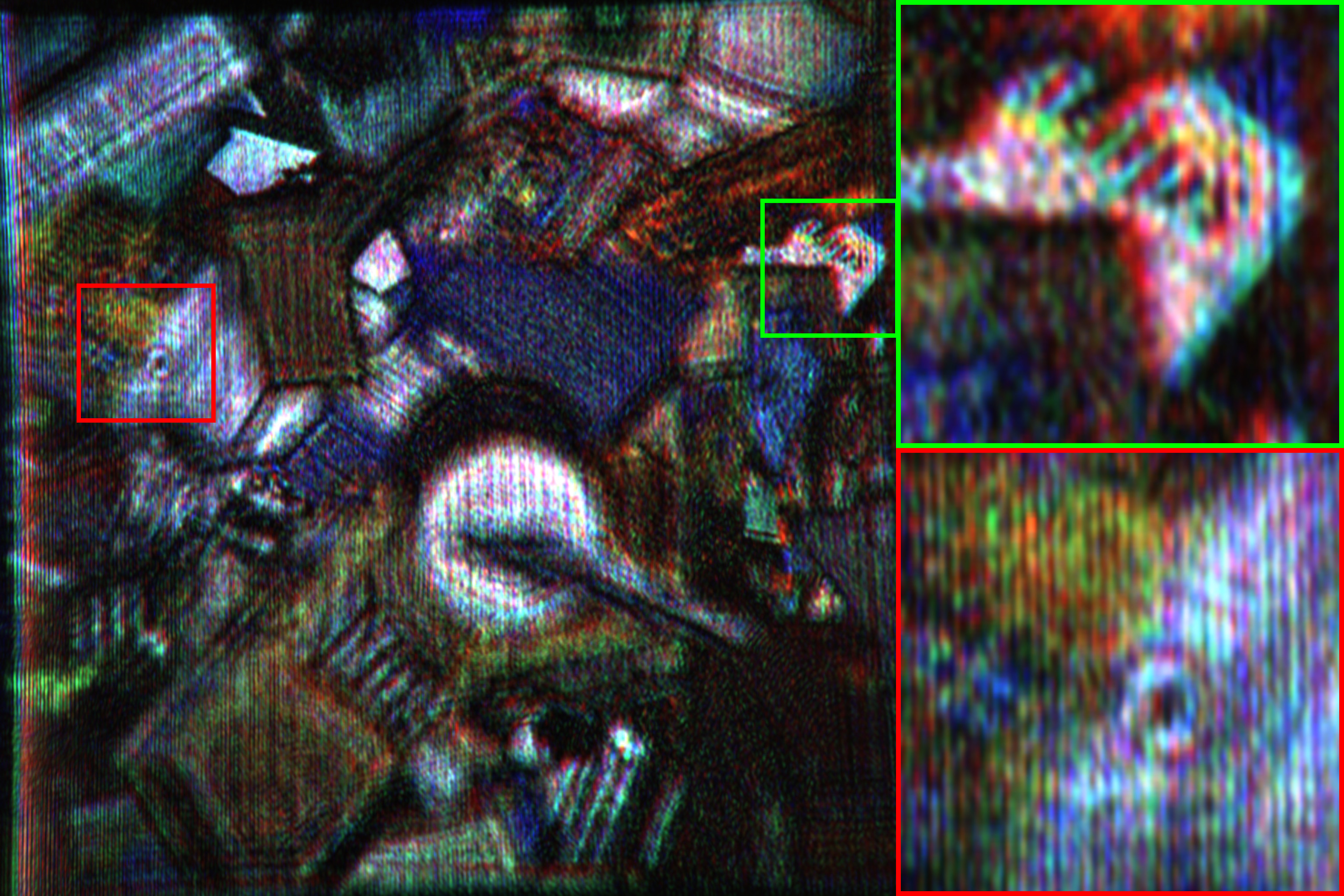}    
    \label{fig:layerbased CGH comparison-projection}
    }
    \caption{Optical reconstructions of holograms generated using different methods, with focus set to front (left) and back (right) planes.}
    \label{fig:optical_recon}
\end{figure*}

For optical reconstruction, we used the May HoloKit equipped with an LCoS IRIS-U62 phase modulator and WikiOptics lasers with wavelengths of 638~nm, 532~nm, and 450~nm.
The phase modulator features a pixel pitch of 3.6~µm and a resolution of $3840 \times 2160$.
To enhance reconstruction quality, a 4f optical system was employed along with double-phase amplitude encoding~\cite{1978dpacRN822} and an off-axis angle of $1.1^\circ$ to encode the complex hologram into a phase-only representation (Fig.~\ref{fig:optical_config}).

Figure~\ref{fig:optical_recon} shows the corresponding optical reconstruction results.
These results are consistent with those observed in numerical reconstructions:
SM-LBM produces pronounced dark outlines around objects, while ADV-LBM and AP-LBM effectively suppress these artifacts.
Notably, AP-LBM achieves superior visual clarity in back-focused regions, demonstrating its robustness in preserving high-frequency details across a large depth range.




\subsection{Quantitative evaluation}
\label{sec:quality_measure}






For 2D holograms, such as those generated by the GS algorithm~\cite{1972GSRN809} or stochastic gradient descent (SGD)-based methods~\cite{2021choiRN258}, metrics like PSNR(Peak Signal-to-Noise Ratio) and SSIM(Structural Similarity Index Map) are appropriate quantitative evaluation tools, as they enable direct comparison between the reconstructed 2D image and its target.
However, in the case of 3D holography, evaluating quality becomes more complex, as it involves comparing a volumetric scene to its holographic reconstruction.

A common approach is to compare a reconstructed focal stack of the hologram with a series of rendered depth images.
However, this method may introduce inaccuracies: when comparing out-of-focus regions (naturally blurred in holographic reconstructions) with sharp rendered images, the resulting metrics may not reliably reflect perceptual quality.

To address this limitation, we employ Focal Image Projection (FIP), an extension of the hologram focal loss proposed in~\cite{2025LiaoRN783}.
Originally designed as a loss function, FIP can also serve as a metric to evaluate hologram quality by comparing only the in-focus regions of reconstructed images with the corresponding rendered image.
FIP generates a composite image by aggregating amplitude information from the focal stack of the hologram reconstruction, using depth-aware masking.
This allows for a direct and meaningful comparison with the rendered image.

FIP is defined as:
\begin{equation}
    \text{FIP}(x,y) = \sum_{i=0}^{n-1} M_i(x,y)\circ|prop.\{H(x,y),-z_i\}|_{A}
    \label{eq:fip}
\end{equation}
Here, $H(x,y)$ is the complex hologram, $|\cdot|_A$ denotes the amplitude component of a complex field, and $M_i(x,y)$ is the depth-aware masking plane, similar to that used in Eq.~\ref{eq:original_SM-LBM}.
The index $i$ runs over the $n$ sampled focal planes.

Since the hologram is generated from the rendered RGB-D image using a layer-based method, the resulting $\text{FIP}(x,y)$ should ideally match the original rendered image $I(x,y)$ if the generation and reconstruction processes are perfect.
Deviations between $\text{FIP}(x,y)$ and $I(x,y)$, measured using PSNR or SSIM, therefore indicate reconstruction error or loss of fidelity.

\begin{figure}
    \centering
    \subfigure[Target image]{
    \includegraphics[width=0.35\linewidth]{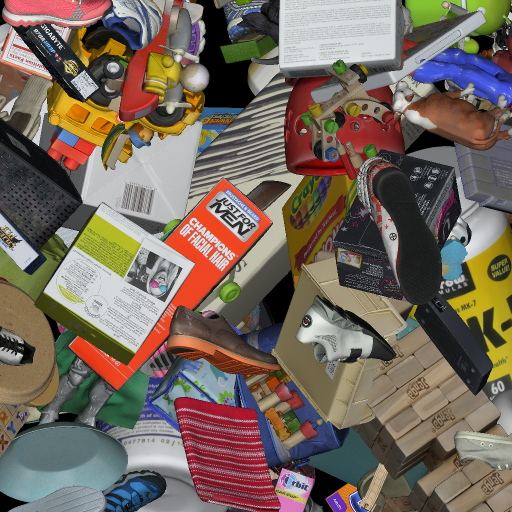}
    \label{fig:layerbased CGH comparison-rgb}
    }
    \subfigure[Original SM-LBM]{
    \includegraphics[width=0.35\linewidth]{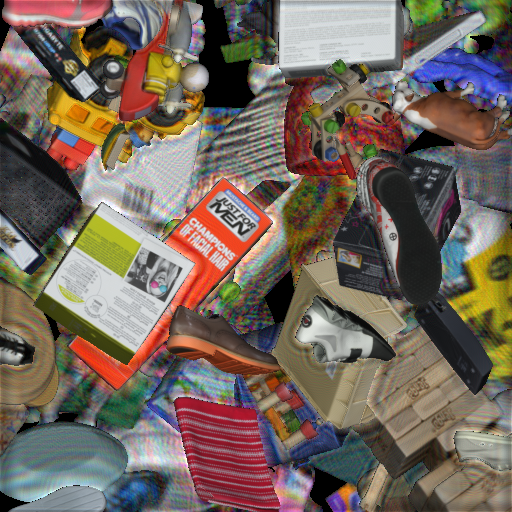}
    \label{fig:layerbased CGH comparison-smlbm}
    }
    \subfigure[ADV-LBM]{
    \includegraphics[width=0.35\linewidth]{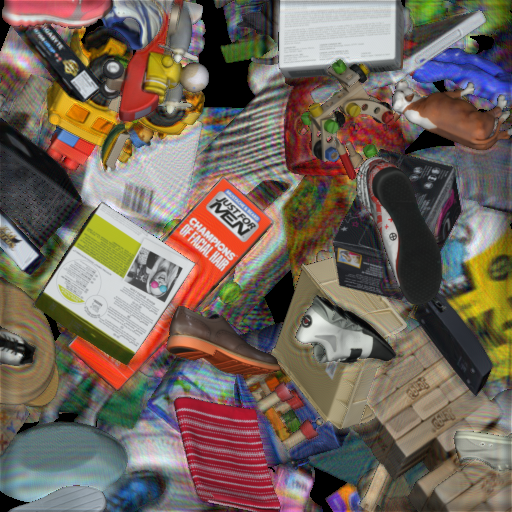}    
    \label{fig:layerbased CGH comparison-Adv}
    }
    \subfigure[AP-LBM]{
    \includegraphics[width=0.35\linewidth]{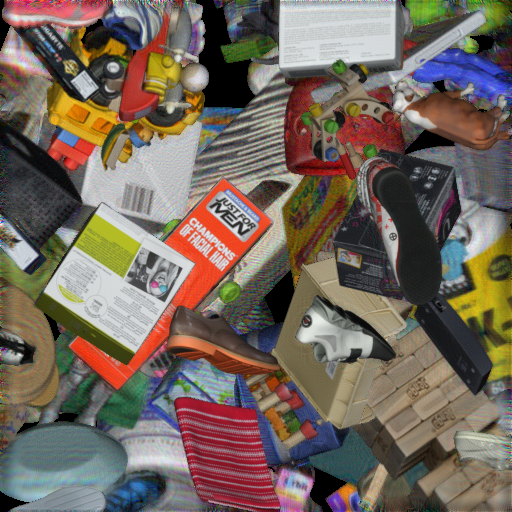}    
    \label{fig:layerbased CGH comparison-projection}
    }
    \subfigure[AP-LBM with artifact reduction]{
    \includegraphics[width=0.35\linewidth]{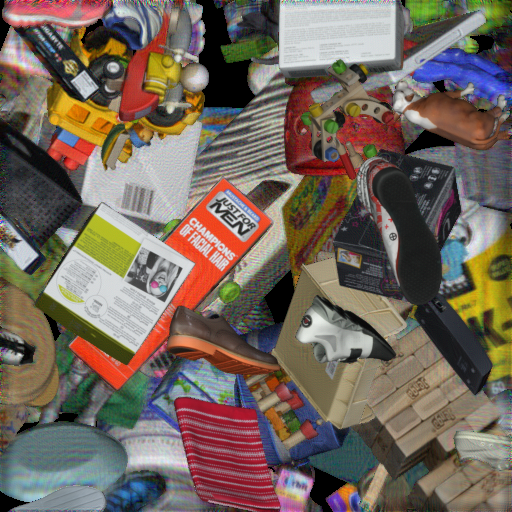}    
    \label{fig:layerbased CGH comparison-Adv}
    }
    \subfigure[AP-LBM with edge padding]{
    \includegraphics[width=0.35\linewidth]{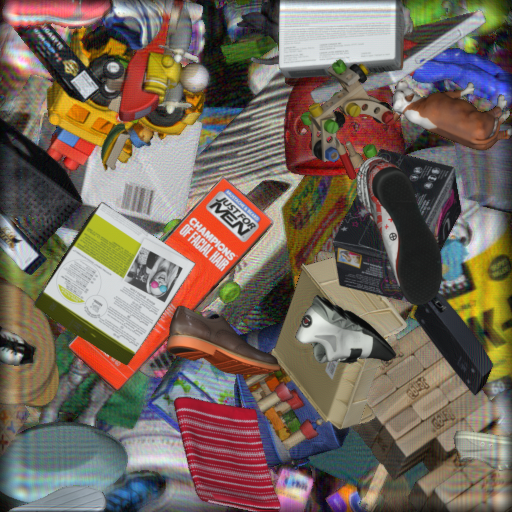}    
    \label{fig:layerbased CGH comparison-projection}
    }
    \caption{(a) Target rendered image used for hologram generation, and focal image projections (FIP) of holograms generated by (b) SM-LBM, (c) ADV-LBM, and (d) AP-LBM.}
    \label{fig:fip_comparison}
\end{figure}

%


We evaluated the PSNR and SSIM between the FIP image and the corresponding rendered image for the three generation methods: SM-LBM, ADV-LBM, and AP-LBM.
For this quantitative comparison, we used 500 holograms from our dataset, each with a resolution of $512 \times 512$.
All other parameters were consistent with those used in the qualitative evaluation setup described in Sec.~\ref{sec:result_qualtitative}.

The quantitative results are summarized in Table~\ref{tab:generation_quality}, and representative FIP images are shown in Fig.~\ref{fig:fip_comparison}.
The proposed method, AP-LBM, achieves the highest performance in terms of both maximum and average PSNR and SSIM values, outperforming the other generation methods.
While ADV-LBM slightly surpasses AP-LBM in the minimum PSNR and SSIM scores, the difference is marginal and not visually significant.

Additionally, we tested AP-LBM in combination with ringing artifact reduction~\cite{2023506ringing} and edge padding, two common techniques for mitigating noise in numerical propagation.
However, both variations resulted in lower reconstruction quality compared to the original AP-LBM.
Further discussion of these artifact reduction techniques is provided in Sec.~\ref{sec:analysis}.

\begin{table}[t]
\centering
\caption{Quantitative evaluation of hologram quality for different generation methods.}
\begin{threeparttable}
\begin{tabular}{c|ccc|ccc}
\hline
\multirow{2}{*}{Generation method} & \multicolumn{3}{c|}{\PSNRFIP} & \multicolumn{3}{c}{\SSIMFIP} \\ \cline{2-7} 
                                   & min      & avg      & max     & min      & avg     & max     \\ \hline
SM-LBM(Eq. \ref{eq:original_SM-LBM})                    & 18.96    & 21.15    & 23.77   & 0.45     & 0.68    & 0.77    \\
ADV-LBM(Eq. \ref{eq:hologram_generation})               & \bf19.77    & 21.96    & 24.62   & \bf0.47     & 0.71    & 0.81    \\
AP-LBM(Eq. \ref{eq:projection})                     & 19.71    & \bf23.99    & \bf27.01   & 0.44     & \bf0.75    & \bf0.87    \\ \hline
Ringing reduction\tnote{*}         & 19.54    & 23.83    & 26.87   & 0.42     & 0.73    & 0.86    \\
Edge-padding\tnote{*}              & 19.69    & 23.19    & 25.99   & 0.45     & 0.73    & 0.84    \\ \hline
\end{tabular}
\begin{tablenotes}
\item[*] AP-LBM with each artifact-reduced numerical propagation method
\end{tablenotes}
\end{threeparttable}
\label{tab:generation_quality}
\end{table}
\section{Analysis and Discussion}
\label{sec:analysis}

In this section, we analyze the influence of key parameters on hologram generation using the proposed AP-LBM method.
We explore how depth range, the number of layers (Sec.~\ref{subsec:depth}), and artifact reduction strategies (Sec.~\ref{subsec:artifact_handing}) affect hologram quality, depth representation, and computational efficiency.

\subsection{Depth range and the number of layers}
\label{subsec:depth}

We analyze how hologram quality is affected by two key configuration parameters: the number of depth layers and the overall depth range of the holographic scene.
To evaluate these effects, we measured the $PSNR_{FIP}$ and $SSIM_{FIP}$ using a test set of 500 holograms, each with a resolution of $512 \times 512$, under various hologram generation configurations.

\begin{table}[h]
\centering
\caption{
Quantitative evaluation of hologram quality under varying depth ranges using AP-LBM at a resolution of $512 \times 512$. 
}
\begin{tabular}{c|ccc|ccc}
\hline
\multirow{2}{*}{Depth-range} & \multicolumn{3}{c|}{\PSNRFIP} & \multicolumn{3}{c}{\SSIMFIP} \\ \cline{2-7} 
                             & min     & avg    & max    & min    & avg    & max    \\ \hline
$1~mm$      & 24.32 & 30.75 & 32.62 & 0.88 & 0.94 & 0.96 \\
$6~mm$      & 22.56 & 27.62 & 30.25 & 0.66 & 0.88 & 0.93 \\
$11~mm$     & 21.03 & 25.90 & 28.84 & 0.54 & 0.82 & 0.91 \\
$16~mm$     & 20.22 & 24.74 & 27.79 & 0.48 & 0.78 & 0.89 \\
$20.334~mm$ & 19.71 & 23.99 & 27.01 & 0.44 & 0.75 & 0.87 \\ \hline
\end{tabular}
\label{tab:depth_range}
\end{table}

For the depth range study, we generated holograms with scene depths of 1 mm, 6 mm, 11 mm, 16 mm, and 20.334 mm.
Table~\ref{tab:depth_range} summarizes the results. At 1 mm, the holograms achieve an average $PSNR_{FIP}$ of 30.75 dB and an $SSIM_{FIP}$ of 0.94, values that are generally indicative of high-quality reconstructions.
At 6 mm, both metrics slightly decrease, though their maximum values remain acceptable. However, beyond 11 mm, the average metrics fall below 30 dB and 0.9, marking noticeable degradation in visual quality.

In particular, minimum $PSNR_{FIP}$ and $SSIM_{FIP}$ values degrade sharply with increasing depth range.
These cases often correspond to scenes where objects are distributed far from the hologram plane.
Despite employing techniques like band-limited ASM and amplitude projection, long propagation distances still introduce significant quality challenges. 
Nonetheless, AP-LBM outperforms previous methods such as SM-LBM and ADV-LBM (Table~\ref{tab:generation_quality}), maintaining usable quality for ML-CGH tasks even at 20.334 mm, achieving up to 27.01 dB $PSNR_{FIP}$ and 0.87 $SSIM_{FIP}$.

It is also worth noting that the \MIT~dataset features a 6 mm depth range. 
When adjusted by the ratio of maximum propagation distance to depth range, this corresponds to approximately 1.6 mm in the \datasetname~configuration. 
This comparison is based on the \MIT~setup using an $8~\mu$m pixel pitch and $384 \times 384$ resolution (max propagation distance: 76.98 mm), versus \datasetname’s 3.6~µm pitch and 20.72 mm propagation range.

\begin{table}[h]
\centering
\caption{
Quantitative evaluation of hologram quality under varying the number of layers using AP-LBM at a resolution of $512 \times 512$.
}
\begin{tabular}{c|ccc|ccc}
\hline
\multirow{2}{*}{\# of layers} & \multicolumn{3}{c|}{\PSNRFIP}  & \multicolumn{3}{c}{\SSIMFIP} \\ \cline{2-7} 
                              & min      & avg      & max      & min      & avg     & max     \\ \hline
10                            & 18.67    & 19.94    & 21.76    & 0.37     & 0.45    & 0.53    \\
100                           & \bf20.11 & 23.11    & 25.11    & \bf0.48  & 0.70    & 0.78    \\
1,000                         & 19.75    & 23.94    & 26.90    & 0.45     & 0.74    & 0.86    \\
10,000                        & 19.71    & \bf23.99 & \bf27.01 & 0.44     & \bf0.75 & \bf0.87 \\ \hline
\end{tabular}
\label{tab:num_of_layers}
\end{table}

In addition to depth range, the number of layers is a critical factor influencing the quality of holograms generated via layer-based methods, as these approaches discretize the continuous depth into multiple focal planes. 
Increasing the number of layers generally improves depth approximation fidelity, but it also introduces a trade-off between reconstruction quality and computational cost.

To evaluate this, we varied the number of layers from 10 to 10,000 and measured the resulting $PSNR_{FIP}$ and $SSIM_{FIP}$.
As shown in Table~\ref{tab:num_of_layers}, significant improvements were observed when increasing the layer count from 10 to 100.
However, the gains became marginal beyond 100 layers.
The configuration with 10,000 layers achieved the highest average and maximum values for both PSNR and SSIM.
Interestingly, the configuration with 100 layers yielded the highest minimum scores, suggesting more consistent quality across diverse scenes.
These findings suggest that although using more than 100 layers generally improves overall quality, it may also introduce occasional degradation in specific cases.




\subsection{Additional notes on artifact handling}
\label{subsec:artifact_handing}

\begin{figure}
    \centering
    \subfigure[ADV-LBM (Eq.~\ref{eq:hologram_generation})]{
    \includegraphics[width=0.35\linewidth]{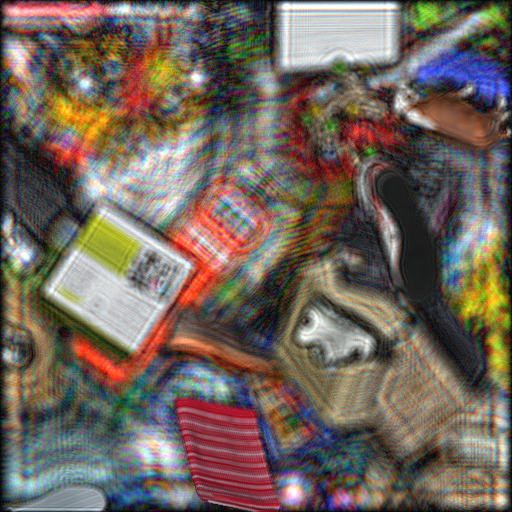}    
    \label{fig:layerbased CGH comparison-SM-LBM}
    }
    \subfigure[AP-LBM (Eq.~\ref{eq:projection})]{
    \includegraphics[width=0.35\linewidth]{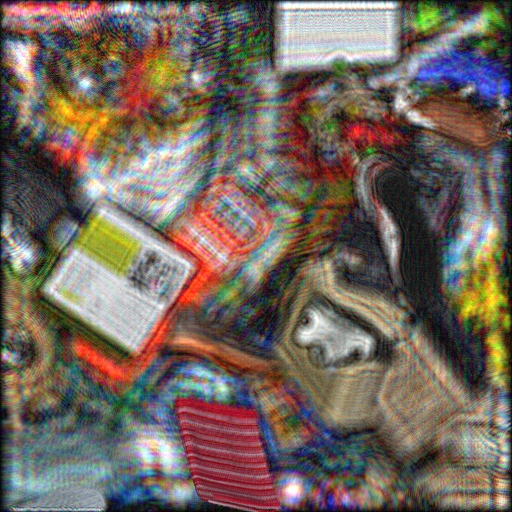}
    \label{fig:layerbased CGH comparison-Adv}
    }
    \subfigure[AP-LBM with ringing artifact reduction]{
    \includegraphics[width=0.35\linewidth]{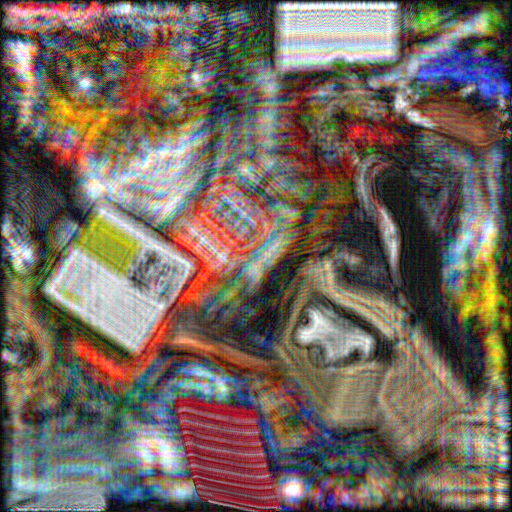}    
    \label{fig:layerbased CGH comparison-projection}
    }
    \subfigure[AP-LBM with edge padding]{
    \includegraphics[width=0.35\linewidth]{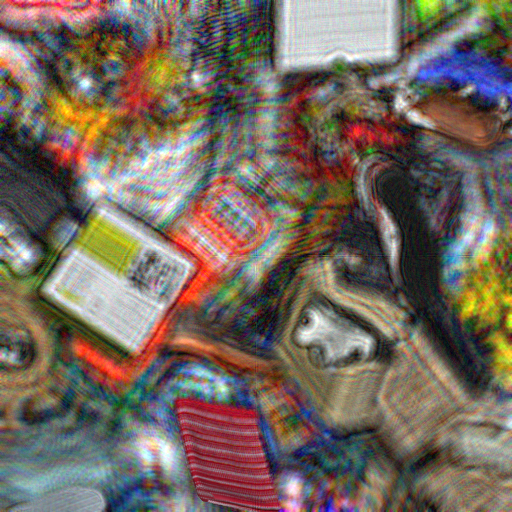}    
    \label{fig:layerbased CGH comparison-SM-LBM}
    }
    \caption{Amplitude maps of complex holograms generated using different methods.
    }
    \label{fig:amplitude}
\end{figure}

\begin{table}[htbp]
\centering
\caption{Hologram quality for AP-LBM with edge-padding under different layer configurations
}
\begin{tabular}{c|ccc|ccc}
\hline
\multirow{2}{*}{\# of layers} & \multicolumn{3}{c|}{\PSNRFIP} & \multicolumn{3}{c}{\SSIMFIP} \\ \cline{2-7} 
                              & min      & avg      & max     & min      & avg     & max     \\ \hline
10                            & 18.50    & 21.22    & 24.45   & \bf0.52     & 0.65    & 0.81    \\
100                           & 19.59    & 22.50    & 25.34   & 0.43     & 0.68    & 0.81    \\
1,000                         & 19.60    & 22.81    & 24.94   & 0.44     & 0.70    & 0.81    \\
10,000                        & \bf19.69    & \bf23.19    & \bf25.99   & 0.45     & \bf0.73    & \bf0.84    \\ \hline
\end{tabular}
\label{tab:num_of_layers_edge_padding}
\end{table}

Although the amplitude projection method improves focal image quality, it may be over-optimized for the angular spectrum method (ASM), leading to increased artifacts as propagation distance grows. As shown in Fig.~\ref{fig:fip_comparison}, FIPs generated with amplitude projection exhibit jitter-like noise near the image borders.

To mitigate these border artifacts, we tested two artifact reduction techniques: \textit{ringing artifact reduction} and \textit{edge padding}.

The ringing artifact primarily arises from zero-padding in ASM~\cite{2023506ringing}.
Although zero-padding is necessary to suppress aliasing in the discrete Fourier transform, it can introduce spurious oscillations (ringing).
A known solution is ringing artifact reduction, which involves dividing the propagated result by a diffraction pattern generated from a uniform white field~\cite{2023506ringing}.
However, in our experiments, this technique offered only marginal improvements, and both PSNR and SSIM were slightly lower compared to AP-LBM without artifact reduction (Table~\ref{tab:generation_quality}).

Edge padding (also known as replicate padding) is another common strategy to reduce edge-related artifacts.
We observed that edge padding can suppress the jittering noise near image borders in AP-LBM and also reduce dark edge artifacts in the hologram amplitude (see Fig.~\ref{fig:amplitude}).
However, this technique introduces inner-frame artifacts in reconstructed images and generally results in lower average PSNR and SSIM compared to zero-padding.
For these reasons, we did not apply artifact reduction techniques in the final version of the \datasetname~dataset.

However, as shown in Table~\ref{tab:num_of_layers_edge_padding}, edge padding provides higher SSIM and PSNR compared with the zero-padding method when holograms are generated using a small number of layers.
Therefore, it remains a practical option in scenarios where layer count is constrained.

\section{Machine Learning Applications}
\label{sec:ml}

To evaluate the applicability of our dataset to machine learning-based CGH (ML-CGH), we trained several models from scratch using the \datasetname~dataset.
These included three hologram generation models, TensorHolography~\cite{2021ShiRN222}, U-Net~\cite{2015UNET}, and Swin-Unet~\cite{2022SWIN810}, as well as the hologram upscaling model H2HSR~\cite{2024NoRN499}.
It is important to note that the primary goal of these experiments is to assess the applicability of the \datasetname~dataset to ML-CGH tasks and to simulate advanced research scenarios, rather than to achieve state-of-the-art performance on the selected models.

We implement each model in PyTorch 2 to ensure that training and testing are performed in the same environment.
The hologram generation models take an RGB image and its corresponding depth map as input of $512\times512$ resolution and output a complex-valued hologram.
The upscaling model takes a low-resolution hologram ($256\times256$ resolution) as input and generates a high-resolution hologram ($512\times512$ resolution) with double the spatial resolution.
For training the generation models, we used 5,000 RGB-D and hologram pairs for training, 500 pairs for validation, and 500 pairs for testing, totaling 6,000 samples from the \datasetname~dataset.
Similarly, for the upscaling model, we used paired low-resolution and high-resolution holograms with the same dataset split.
\begin{table}[htbp]
\centering
\setlength{\tabcolsep}{5pt}
\caption{Specifications and hologram reconstruction performance of ML-CGH models trained from scratch using the \datasetname~dataset.
PSNR and SSIM are reported for amplitude, numerical reconstruction, and FIP.
Latency refers to the average inference time per sample, measured in milliseconds (ms) on a single NVIDIA RTX A6000~GPU.
}
\label{tab:machine_learning}
\begin{tabular}{c|c|c|ccc|ccc}
\hline
\multirow{2}{*}{Model} &
  \multirow{2}{*}{Params} &
  \multirow{2}{*}{Latency(ms)} &
  \multicolumn{3}{c|}{PSNR(dB)} &
  \multicolumn{3}{c}{SSIM} \\ \cline{4-9} 
                                           &      &     & Amp.    & Recon.  & FIP     & Amp.    & Recon.  & FIP     \\ \hline
TensorHolography~\cite{2021ShiRN222}                    & 149K & 3.3 & 24.3    & 25.3    & 20.7    & 0.62    & 0.71    & 0.60    \\
Unet~\cite{2015UNET}                        & 13M  & 2.8 & 26.4    & 28.1    & 21.7    & 0.70    & 0.79    & 0.64    \\
Swin-Unet~\cite{2022SWIN810}                & 42M  & 165 & \bf26.8 & \bf28.7 & \bf21.8 & \bf0.72 & \bf0.80 & \bf0.65 \\ \hline
H2HSR\textsubscript{Swin}~\cite{2024NoRN499} &
  12M &
  640 &
  \bf26.3 &
  \bf27.2 &
  \bf21.5 &
  \bf0.7 &
  \bf0.76 &
  \bf0.63 \\
H2HSR\textsubscript{RDN}\cite{2024NoRN499} & 22M  & 148 & 25.3    & 25.7    & 21.0    & 0.66    & 0.72    & 0.60    \\ \hline
\end{tabular}

\end{table}
\begin{figure}
    \centering
    \subfigure[GT]{
    \includegraphics[width=1.0\linewidth]{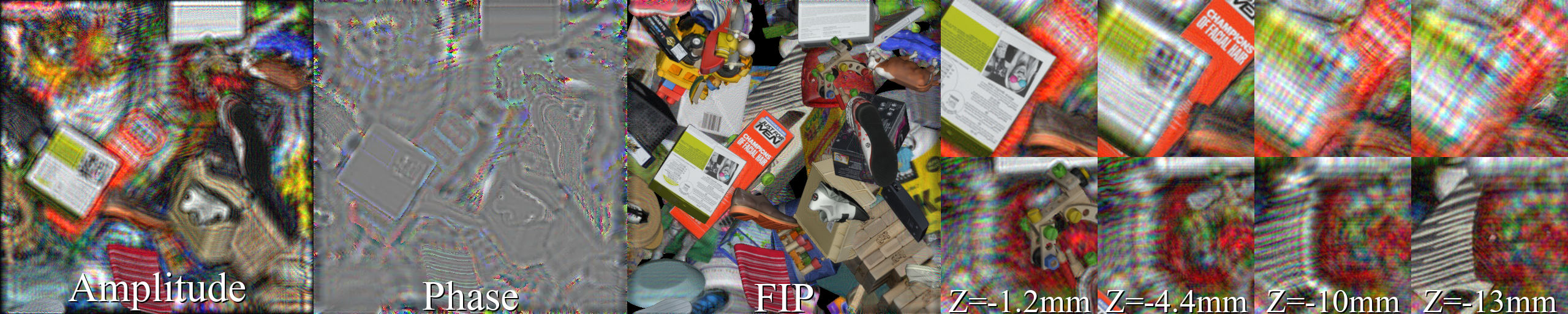}
    }
    \subfigure[TensorHolography]{
    \includegraphics[width=1.0\linewidth]{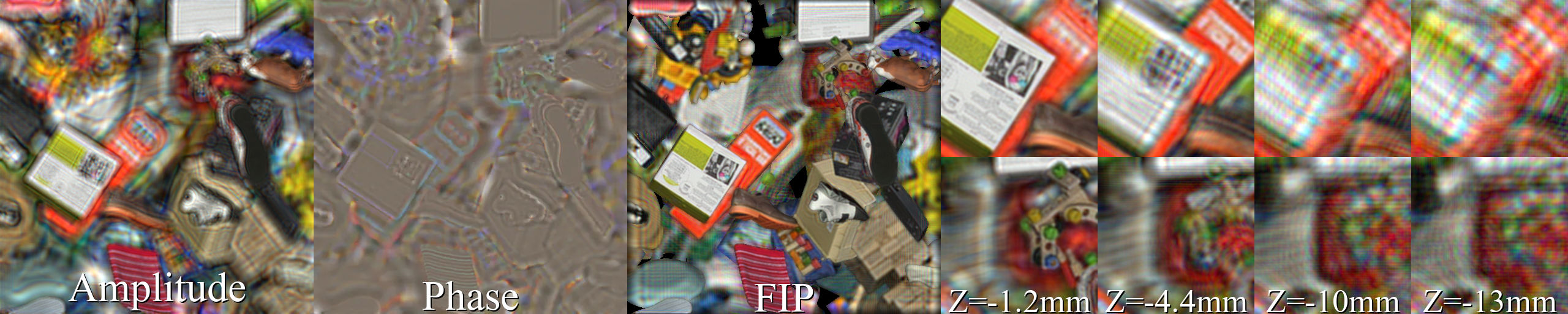}
    }
    \subfigure[U-Net]{
    \includegraphics[width=1.0\linewidth]{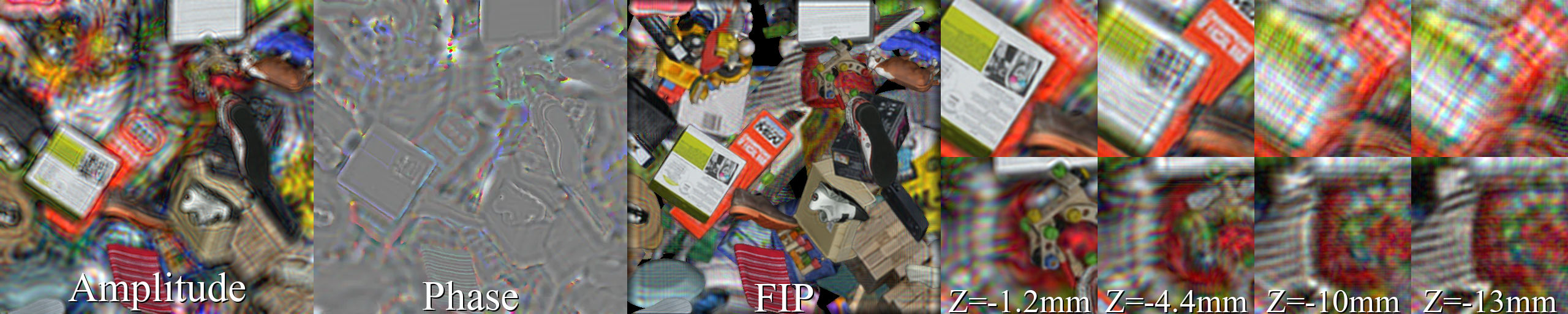}
    }
    \subfigure[Swin-Unet]{
    \includegraphics[width=1.0\linewidth]{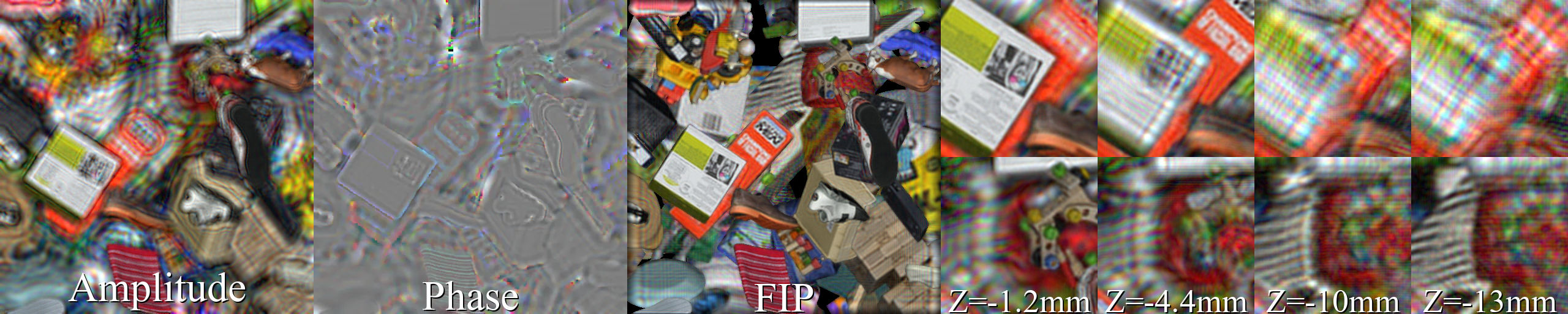}
    }
    \caption{
    Qualitative comparison of ML-CGH results. Reconstructed focal planes are located from -1.2 mm (front focus) to -13 mm (back focus). The hologram generation models (TensorHolography, U-Net, and Swin-Unet) take an RGB-D image as input and generate a $512 \times 512$ hologram. 
    }
    \label{fig:ML_result}
\end{figure}

\begin{figure}
    \centering

    \subfigure[$H2HSR_{Swin}$]{
    \includegraphics[width=1.0\linewidth]{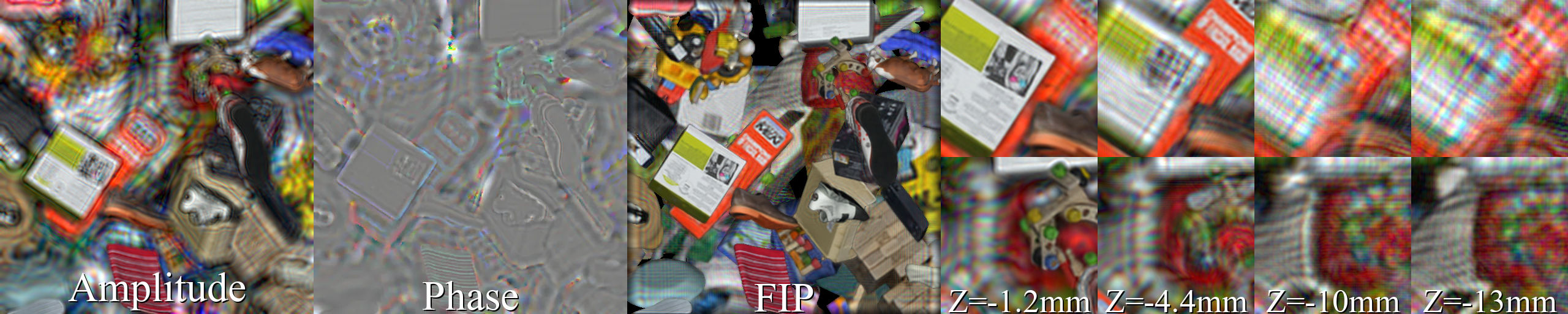}
    }
    \subfigure[$H2HSR_{RDN}$]{
    \includegraphics[width=1.0\linewidth]{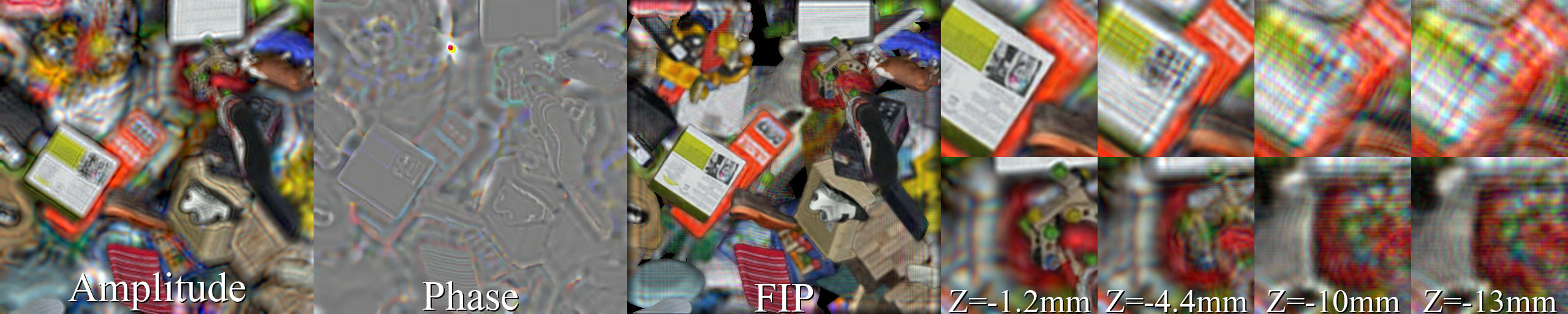}
    }
    
    \caption{
    Qualitative comparison of ML-CGH results. Reconstructed focal planes are located from -1.2 mm (front focus) to -13 mm (back focus). 
    The upscaling models ($H2HSR_{Swin}$ and $H2HSR_{RDN}$) take a $256 \times 256$ hologram as input and produce a $512 \times 512$ output.
    }
    \label{fig:h2hsr_result}
\end{figure}

Table~\ref{tab:machine_learning} summarizes the quantitative results in terms of PSNR and SSIM, measured for amplitude, reconstructed images, and focal image projection (FIP).
Fig.~\ref{fig:ML_result} and Fig.~\ref{fig:h2hsr_result} present representative visual outputs from each model.
All models achieved reasonable performance, demonstrating that the \datasetname~dataset is well-suited for training both hologram generation and upscaling models.

For the hologram generation models, we observed that as model complexity (i.e., the number of parameters) increased, the quality of the generated holograms also improved when trained on the \datasetname~dataset.
Interestingly, when trained on the \MIT~dataset, TensorHolography achieved the best performance among all models, despite having fewer parameters than the others.
We attribute this difference to the distinct hologram configurations in the two datasets, particularly in terms of depth range and generation method.
The \datasetname~dataset features a substantially larger depth range and employs a layer-based generation method, whereas \MIT~uses a point-based method, which likely affects the optimal network architecture for each case.
We believe that \datasetname~presents a more challenging and realistic configuration (e.g., larger depth range and complex occlusions), which encourages the development of more robust and generalizable ML-CGH models.

In the upscaling model experiments, H2HSR\textsubscript{Swin} achieved higher PSNR and SSIM scores than H2HSR\textsubscript{RDN}, despite having fewer parameters.
However, its actual computational cost was higher, likely due to the architectural complexity of the Swin Transformer.
Nonetheless, the performance of both models demonstrates that the upscaling task can be effectively trained using the \datasetname~dataset.

These results highlight the practical value of the \datasetname~dataset for ML-CGH research and development.
Looking ahead, we believe that the dataset can enable future work in areas such as AR/VR holographic displays, depth-aware visualization, and real-time holographic rendering pipelines.

\section{Conclusion and Future Work}


We introduced \datasetname, a publicly available hologram dataset designed to ML-CGH across large depth ranges and high spatial resolutions.
The dataset comprises 6,000 RGB-D and complex hologram pairs rendered at multiple resolutions ($256 \times 256$ to $2048 \times 2048$), making it suitable for a wide range of ML-CGH tasks, including hologram generation and upscaling.
To enable high-quality hologram generation over extended depth ranges, we proposed a novel amplitude-projection-enhanced layer-based method (AP-LBM), which improves reconstruction fidelity by projecting amplitude information at corresponding depth planes.
We also introduced a quantitative evaluation method, focal image projection (FIP), to assess hologram accuracy against rendered RGB-D images. 
Experimental results show that AP-LBM achieves the highest $PSNR_{FIP}$ and $SSIM_{FIP}$ scores among evaluated methods, confirming its effectiveness.
In addition, we conducted an in-depth parameter analysis, revealing how depth range and layer count affect hologram quality.
These findings provide practical guidance for balancing reconstruction quality and computational efficiency when generating large-scale hologram datasets.

\paragraph{Limitations and Future Directions.}
Despite the improvements introduced by AP LBM, the method remains susceptible to limitations such as ringing artifacts and unnatural blur in defocused regions. The unnatural blur effect stems from smooth phase initialization, which, while reducing speckle noise in focused areas, can make defocus transitions appear less natural.
Random phase initialization could mitigate this, but it may also introduce speckle noise in focused regions.
Recent hologram generation methods, such as LDI~\cite{2022Shi261} and Gaussian Splatting~\cite{gaussianChoi2025}, demonstrate high-quality holographic reconstructions.
However, they are not free from ringing artifacts and unnatural blur, as they also rely on the angular spectrum method with smooth phase initialization.
Moreover, these methods require significant input pre-processing, volumetric rendering for LDI and light field generation for Gaussian Splatting, whereas RGB-D images can be rendered directly from 3D mesh data with minimal overhead.
Layer-based methods also have limitations in reproducing accurate parallax, particularly at extreme viewing angles.
Nonetheless, they remain highly practical due to their simple generation process and strong front-view reconstruction quality.

In future work, we plan to explore hybrid generation techniques that combine the interpretability and simplicity of layer-based methods with the expressiveness of point- and volume-based representations.
We also aim to expand the dataset to include dynamic scenes and broader variations in lighting and material properties, further enhancing its applicability to real-world holographic display systems.


\section{Conflict of Interest}
The authors declare that they have no conflicts of interest.

\section{Acknowledgements}
This work was supported by the National Research Foundation of Korea (NRF) through the Ministry of Education's Basic Science Research Program (Grant 2021R1I1A3048263, 50\%) and by the Institute of Information and Communications Technology Planning and Evaluation (IITP) grant funded by the Korea Government (MSIT) (Grant 2019-0-00001, 50\%).

\section{Data Availability Statement}
The dataset, \datasetname, is publicly available in the Hugging Face Datasets repository at https://hpc-lab-koreatech.github.io/KOREATECH-CGH/

\bibliographystyle{IEEEtran}
\bibliography{bib.bib}

\end{document}